\documentclass[runningheads]{llncs}

\usepackage{eccv}

\usepackage{eccvabbrv}
\definecolor{lowred}{RGB}{238,18,137}
\def\mypar#1{\vspace{1mm}{\noindent\bf #1.}\hspace{1mm}}
\newcommand{\methodname}{Wat3R}
\newcommand{\datasetname}{Water3D}

\newcommand{\dplustem}[1]{\fontsize{5pt}{0.3em}\selectfont (\textbf{\textcolor{lowred}{#1}})}
\newcommand{\dplus}[1]{\raisebox{-1.05ex}{\fontsize{5pt}{0.3em}\selectfont (\textbf{\textcolor{lowred}{#1}})}}
\usepackage[colorinlistoftodos]{todonotes}
\usepackage{multirow}
\usepackage{booktabs}
\usepackage{makecell}
\usepackage{graphicx}
\usepackage[table]{xcolor}
\usepackage{xcolor}
\usepackage{adjustbox}

\usepackage{amssymb}
\usepackage{caption}
\usepackage{gensymb}
\usepackage{booktabs}
\usepackage{float}
\usepackage[accsupp]{axessibility}  

\usepackage[colorlinks,
            linkcolor=red,
            ]{hyperref}

\usepackage{orcidlink}

\begin{document}

\title{Wat3R: Underwater 3D Geometry Learning without Annotations}
\titlerunning{Wat3R}

\author{Jiangwei Ren  \and Xingyu Jiang\inst{\dag} \and Zijie Song  \and Wei Xu  \and Hongkai Lin \and \\ Dingkang Liang  \and Xiang Bai }
\authorrunning{J.~Ren et al.}
\institute{Huazhong University of Science and Technology\\
\email{\{jwren,jiangxy998,dkliang,xbai\}@hust.edu.cn}}

\maketitle
\let\thefootnote\relax\footnotetext{$\dag$ Corresponding author.}

\begin{abstract}

Estimating 3D geometry in underwater environments presents unique challenges due to light attenuation, scattering, and the absence of large-scale, high-quality 3D annotations. Pioneering methods rely on massive dense annotations that are impractical in underwater settings. In this paper, we propose \textbf{\methodname{}}, a cross-domain semi-supervised learning framework designed to adapt feed-forward 3D reconstruction models from air to underwater scenes. Uniquely, our method eliminates the need for any annotated underwater data following a teacher-student architecture, that learns robust geometry representations merely on abundant unlabeled real underwater video footage. We also design a cross-view consistency loss that leverages geometric cues from other views to compensate for the information degradation in the current view caused by water attenuation and scattering.
Furthermore, considering the lack of comprehensive evaluation benchmarks, we construct \textbf{\datasetname{}}, a diverse dataset covering various water bodies and underwater scenarios, designed for geometric task evaluation.
Experimental results demonstrate that \methodname{} outperforms current state-of-the-art methods in underwater multi-view depth estimation and point cloud reconstruction. The dataset and code are available at \url{https://github.com/LSXI7/Wat3R} .

\keywords{Underwater Vision, Geometry Estimation, VGGT, Cross-domain Semi-supervised Learning}
\end{abstract}

\section{Introduction}

Underwater visual geometry estimation aims to recover the 3D structures, including camera poses, depths, and point clouds, from multi-view underwater imagery.
This capability underpins practical applications ranging from underwater robot navigation and obstacle avoidance, to marine mapping, terrain modeling, and underwater archaeology~\cite{kim2009pose,liu2025underwater}. 
Unlike on-land scenes, underwater environments present unique challenges caused by the light absorption and scattering, as well as view-dependent degradation. These physical difficulties further lead to the critical scarcity of large-scale, high-quality 3D annotations~\cite{randall2023flsea} in this domain, posing great challenges in training a well-performing model.

In recent years, 3D vision has witnessed a paradigm shift from classical multi-view geometry pipelines to feed-forward neural reconstruction models. Advanced methods like  DUSt3R~\cite{wang2024dust3r}, VGGT~\cite{wang2025vggt} and their successors~\cite{leroy2024grounding,yang2025fast3r,wang20253d,cabon2025must3r} have demonstrated remarkable performance in recovering 3D geometry in a single forward pass. These models learn strong geometric priors from massive on-land datasets with dense 3D ground truths, \emph{a.k.a.} camera poses and depths. However, directly deploying these powerful models in underwater scenarios results in poor generalization due to the significant domain shift. And the difficulty in obtaining dense 3D labels for real underwater data also prevents the direct training of effective, generalizable models tailored to underwater environments.

\begin{figure}[t]
  \centering
  \includegraphics[width=0.95\linewidth]{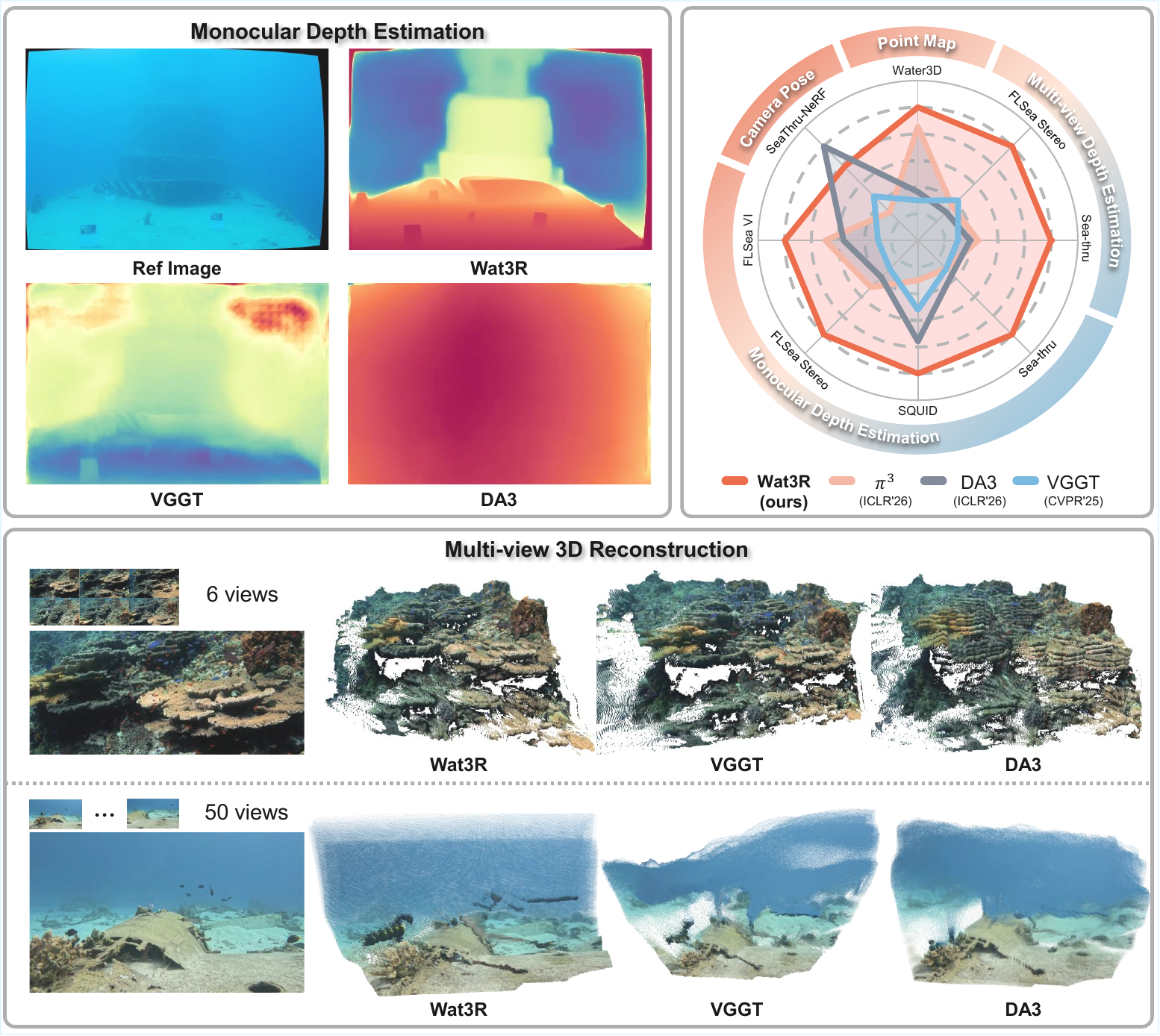}
    \vspace{-2mm}
  \caption{\textbf{\methodname{}} reconstructs from the open-domain underwater images in a feed-forward manner without requiring any underwater 3D annotations.  Our Wat3R achieves significant enhancement in both single-view and multi-view tasks. Statistic results also reveal the superior performance of our \methodname{} against the SOTA.}
  \label{fig:intro}
\end{figure}

\begin{figure*}[t]
\centering
\makebox[\textwidth][c]{%
\begin{minipage}[t]{0.58\textwidth}
\vspace{0pt}
\centering
\includegraphics[width=\linewidth]{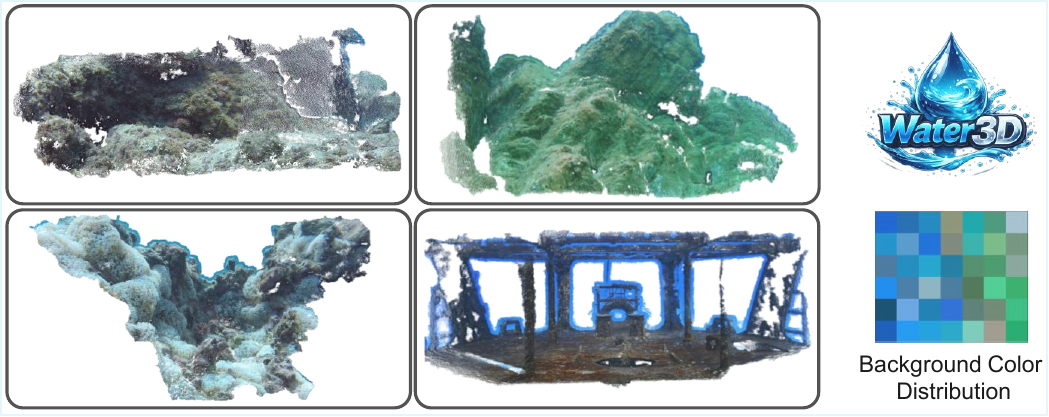}
\end{minipage}%
\hspace{0.02\textwidth}%
\begin{minipage}[t]{0.38\textwidth}
\vspace{5pt}
\centering
\footnotesize
\renewcommand{\arraystretch}{1.08}
\begin{adjustbox}{width=\linewidth}
\begin{tabular}{lccc}
\toprule
\multirow{3}{*}{\vspace{+0.5em}\textbf{Dataset}}
& \multirow{3}{*}{\vspace{+0.5em}\textbf{\#Scenes}}
& 
\multicolumn{2}{c}{\textbf{Annotations}} \\
\cmidrule(lr){3-4}
& & 

Depth & Pose \\
\midrule
SeaThru~\cite{akkaynak2019sea} & 5 & \checkmark & --\\
SQUID~\cite{berman2020underwater} & 4 & \checkmark & --\\
FLSea VI~\cite{randall2023flsea} & 12 & \checkmark & --\\
FLSea Stereo~\cite{randall2023flsea} & 5 & \checkmark & --\\
SeaThru-NeRF~\cite{levy2023seathru} & 4 & -- & \checkmark\\
\textbf{Water3D (Ours)} & \textbf{42} & \textbf{\checkmark} & \textbf{\checkmark}\\
\bottomrule
\end{tabular}
\end{adjustbox}
\end{minipage}%
}
\vspace{-2mm}
\caption{
\textbf{Overview of our constructed \textbf{\datasetname{}} dataset.}
Left: point cloud visualization and background color distribution of {\datasetname{}}.
Right: comparison of existing underwater datasets in terms of scene scale and available annotations.
Our {\datasetname{}} consists of various underwater conditions with both depth and pose annotations.
}

\label{fig:dataset_comparison}
\end{figure*}

To relieve the above dilemma, we propose \textbf{Wat3R}, a semi-supervised framework that adapts VGGT to the underwater domain without any underwater 3D annotations. As shown in \cref{fig:intro}, our method enables feed-forward reconstruction across underwater scenes with complex water conditions. Following a teacher-student learning paradigm, we first simulate various underwater degradations on existing annotated on-land datasets (the original training set of VGGT) as the required labeled data, which helps to initialize strong geometric priors for underwater environments. To further refine and generalize the model to a real underwater domain, we also collect a large amount of real underwater video footage as the unlabeled training set. Besides, to address the visual degradation caused by water attenuation and scattering, we introduce the \textit{Cross-view Consistent Loss}, which integrates geometric cues from other views to compensate for the information degradation in the current view. Considering the incomplete annotations and insufficient coverage of existing underwater benchmarks, we construct \textbf{Water3D}, which contains a wide spectrum of underwater conditions and scenarios with correct camera poses and depths as shown in~\cref{fig:dataset_comparison}.
    Extensive experiments on public datasets and our Water3D demonstrate that our Wat3R significantly outperforms strong baselines and recent feed-forward alternatives, delivering robust performance even in poor visibility underwater regions.

In summary, our contributions are as follows:
\begin{itemize}
    \item We propose \textbf{Wat3R}, the first semi-supervised VGGT-based framework that generalizes to diverse underwater scenes without requiring underwater 3D annotations.
    It offers a practical paradigm for geometric learning in other complex environments, just leveraging unlabeled real videos.
    \item We introduce a \textit{Cross-View Consistent Loss}, which makes it easy to learn geometry cues from heavily degraded regions by aggregating the information from other related views as compensation.
   \item We construct \textbf{\datasetname{}}, an underwater multi-view dataset with comprehensive 3D annotations. It contains diverse underwater conditions.
\end{itemize}

\section{Related Work}

\subsection{Feed-forward 3D Reconstruction Models}

Classical 3D reconstruction is built on multi-view geometry and optimization-based pipelines. Wherein the camera poses and scene structure are recovered through feature matching, epipolar geometry, and global refinement~\cite{hartley2003multiple,snavely2006photo,agarwal2011building,schonberger2016structure,crandall2011discrete,crandall2012sfm,cui2017hsfm}.
These systems achieve high accuracy and scalability. 
However, they rely on complex multi-stage pipelines and iterative optimization, leading to high computational cost and limited robustness in challenging imaging conditions like underwater scenes.
Recent work reforms 3D reconstruction toward feed-forward inference. 
DUSt3R~\cite{wang2024dust3r} initiates this paradigm by directly regressing dense geometry from image pairs without known camera poses. MASt3R~\cite{leroy2024grounding} strengthens this 3D prior for correspondence and alignment. 
Subsequent methods~\cite{cabon2025must3r,yang2025fast3r,wang20253d,wang2025pi,maggio2025vggt} progressively extend this framework toward multi-view inference, higher efficiency, and SLAM-oriented applications. 
VGGT~\cite{wang2025vggt} further improves reconstruction accuracy through large-scale training and multi-task joint optimization. 
More recently, MapAnything~\cite{keetha2025mapanything} expands the input space by incorporating multiple geometric inputs, while Depth Anything 3(DA3)~\cite{lin2025depth}  advances large-scale generalization with unified depth–ray modeling.
However, the success of these models critically depends on massive labeled 3D datasets. 
In underwater scenes, obtaining those geometric annotations is difficult due to the poor visibility and unknown underwater imaging, which fundamentally limits the applicability of data-hungry reconstruction models.

\subsection{Underwater Vision and Geometry Perception}

Underwater visual perception is strongly influenced by the physical image formation process, where light attenuation and scattering significantly degrade visual signals. A large body of work~\cite{akkaynak2019sea,huang2023contrastive,liu2025toward} has focused on underwater image restoration and enhancement, aiming to remove color distortion and scattering effects to improve visual quality. Physically grounded underwater imaging models~\cite {akkaynak2018revised} therefore play a central role in characterising underwater visual environments. Several reconstruction and perception frameworks~\cite{yu2022udepth,levy2023seathru,wang2024underwater} explicitly incorporate these models to jointly reason about scene geometry and underwater light transport. Recent work~\cite{xu2025nautilus} further shows that incorporating underwater imaging priors can improve LMMs for the understanding of underwater scenes. 

Due to the scarcity of large-scale underwater datasets with accurate 3D annotations, several studies explore synthetic data generation using physics-based rendering engines~\cite{lv2025uwstereo,wu2025stereoadapter} or generative models~\cite{zhang2024atlantis,lin2025unified}. However, these approaches are typically designed for stereo or monocular perception tasks and often exhibit domain discrepancies when transferred to real underwater environments.  NeRF~\cite{levy2023seathru,tang2024neural,zhou2025waterhe} and 3D Gaussian Splatting methods~\cite{li2025watersplatting,yang2025seasplat,wang2025uw} have also been applied to underwater scene reconstruction. These methods benefit from the compatibility between volumetric rendering formulations and underwater imaging models, enabling joint modeling of scene geometry and light transport. Nevertheless, they generally rely on known camera poses or sparse geometric priors and require optimization-based reconstruction.
In contrast, our work focuses on feed-forward multi-view geometry learning in underwater scenes without requiring underwater 3D annotations. For this purpose,  we introduce a cross-domain semi-supervised learning framework together with a cross-view consistency loss to address underwater degradation for better training.

\section{Method}

This paper aims to recover camera poses, depths and point clouds from collected underwater views in a single feed-forward pass. 
To achieve this, several key challenges we may face: \textit{(i) How to train a model under limited or no 3D annotations for underwater scenes. (ii) How to restrain the information loss caused by unknown underwater degradation.} 
 In the following, we will alleviate these concerns by applying cross-domain semi-supervised learning onto the advanced pretrained visual geometry model. Specifically, we first introduce the problem definition and notation of the base model VGGT~\cite{wang2025vggt}, then describe the overall framework and training strategy, and finally detail our consistency-aware losses.

\begin{figure*}[tp]
  \centering
  \includegraphics[width=0.95\linewidth]{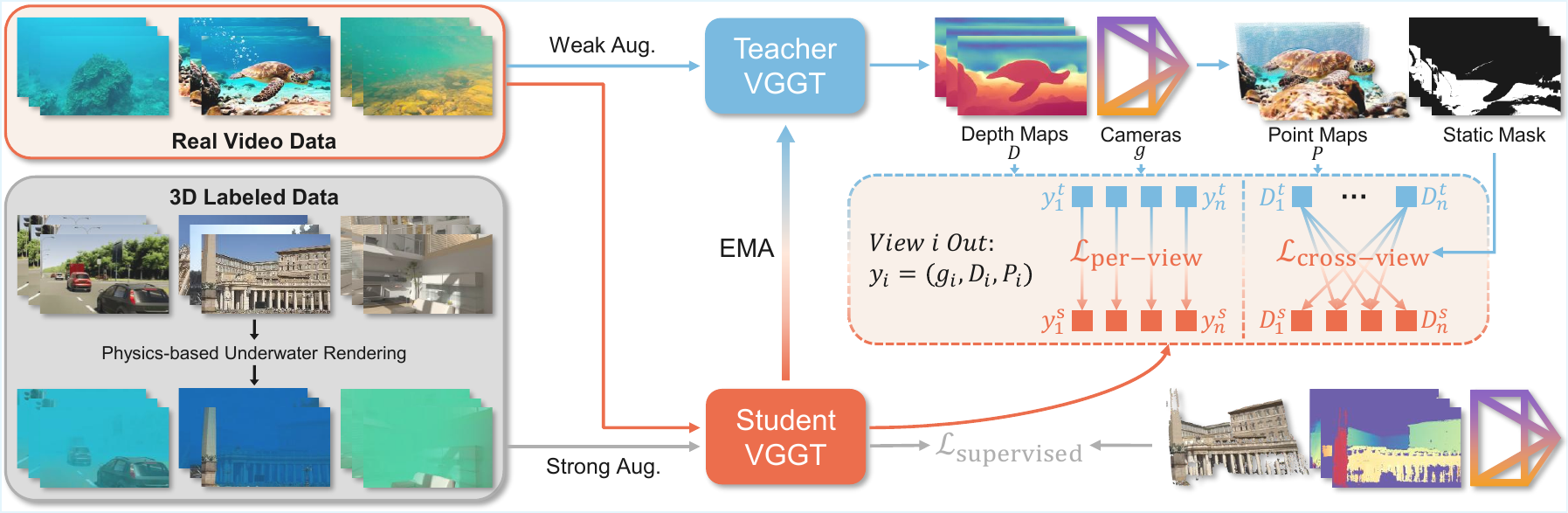}
  \caption{\textbf{Overview of our \methodname{} framework.}
Our pipeline follows a Mean Teacher semi-supervised paradigm, where the teacher network produces pseudo-labels for depth, camera parameters, and point maps to supervise the student network.
Training leverages labeled synthetic underwater data together with unlabeled real underwater videos, enabling adaptation without underwater 3D annotations.
Additional per-view and cross-view consistency losses enforce multi-view geometric coherence. }
\label{fig:framework}  
\end{figure*}

\subsection{Problem Definition and Notation}
\label{sec:problen}

Given a sequence of underwater RGB frames $(I_i)_{i=1}^{N}$, the target of our model is to predict a set of geometric attributes 
$y_i$ for each view $i$, formulated as:
\begin{equation}
(y_i)_{i=1}^{N}
=
f\!\left((I_i)_{i=1}^{N}\right),
\end{equation}
where $y_i = (g_i, D_i, P_i)$ denotes the predicted geometric attributes of view $i$ corresponding to the camera parameters, depth map, and point map, respectively. 
$g_i\in\mathbb{R}^{9}$ encodes the camera pose between view $i$ and the first camera,  containing the rotation quaternion, translation vector, and field of view.
$D_i\in\mathbb{R}^{H\times W}$ denotes the depth map; 
and $P_i\in\mathbb{R}^{3\times H\times W}$ is the predicted 3D point map expressed in the coordinate frame of the first camera. 
 Following this feed-forward framework, VGGT~\cite{wang2025vggt} is known as the pioneering work, which is trained on large-scale 3D-annotated on-land datasets and has shown remarkable performance. Therefore, we chose VGGT as the base model such that it can provide strong geometric priors when adapting to underwater environments. However, it is impractical to directly train VGGT for underwater adaptation due to the lack of 3D annotated training sets.

\subsection{Cross-domain Semi-supervised Learning}
\label{sec:architecture}
Considering the lack of underwater 3D annotations, we investigate a semi-supervised learning framework that adapts a pretrained multi-view geometry model from land to underwater environments. Our key idea is to transfer geometric priors to underwater scenes through consistency-driven teacher–student training. As shown in Fig.~\ref{fig:framework}, the framework follows a Mean Teacher~\cite{tarvainen2017mean} design, where the teacher provides stable pseudo geometric supervision while the student learns from both labeled and unlabeled data.

To be specific, we first simulate various underwater degradations on existing annotated on-land datasets (the original training set of VGGT) as the required labeled data, which helps to initialize strong geometric priors for underwater environments. We then collect a large amount of real underwater video footage as the unlabeled training set, that helps to further refine and generalize the model to the real underwater domain. 
During training, the student network is updated through backpropagation, while the teacher follows an exponential moving average (EMA) of the student’s parameters with the update rule:
\begin{equation}
\theta_t^k = (1-\lambda)\cdot \theta_t^{k-1} + \lambda\cdot \theta_s^k,
\label{eq:ema}
\end{equation}
where $\theta^k_s$ and $\theta^k_t$ are the student parameters and teacher parameters in iteration $k$, respectively. $\lambda$ is the smoothing coefficient.

\mypar{Synthesized Underwater Data (Labeled)}
We generate underwater-style training images using a simplified version of the revised underwater imaging model~\cite{akkaynak2018revised}. 
The image formation is modeled as:

\begin{equation}
I = J e^{-\beta^{D} z} + B^{\infty}(1-e^{-\beta^{B} z}),
\end{equation}
where $J$ and $I$ denote the clear image and its underwater degradation and $z$ is the depth of the scene. $B^\infty \in \mathbb{R}^3$ denotes the background light, while 
$\beta^D \in \mathbb{R}^3$ and $\beta^B \in \mathbb{R}^3$ model the attenuation of direct transmission and backscatter, respectively. 
All three parameters are randomly sampled within $[0,1]$.
The multi-view datasets often contain incomplete or noisy depth annotations, which can create partial water-rendering artifacts when used directly in the image formation model.
Therefore, we estimate dense depth maps using the monocular depth model DA3MONO-LARGE~\cite{lin2025depth} for rendering the underwater appearance, while keeping the original geometric annotations as the supervised targets.
To control the visible range of each scene, the depth maps are linearly rescaled to a random physical range. 
The attenuation coefficients are sampled while enforcing the physical constraint that red attenuates faster than green and blue~\cite{akkaynak2019sea}. 
To simulate spatially varying attenuation, the sampled coefficients are modulated by a smooth random spatial map obtained by repeatedly smoothing Gaussian noise. Several visual results of synthetic underwater can be seen in the bottom left of Fig.~\ref{fig:framework}, which shows the good simulation of different underwater degradations.

\mypar{Real Underwater Video Data (Unlabeled)}
To capture the diverse visual conditions of real underwater environments, we construct a large-scale video collection combining a curated scientific dataset with in-the-wild footage. 
We collect around $10,000$ raw underwater videos from public sources, but only 5,504 clips are left after manual filtering. 
We retain clips with continuous shots, visible static structures, reasonable camera motion, sufficient view overlap, and usable image quality.
We remove surface or aerial videos, edited clips with frequent cuts, near duplicates, severe blur or compression, dominant overlays, and clips dominated by open water or moving objects. 
Frames are sampled every ten frames to reduce temporal redundancy, yielding $359k$ training images in total.

\mypar{Sequence-level Training Augmentation}
As a semi-supervised learning framework, data augmentation is critical for stable training.  Following single image based  task~\cite{zhao2023augmentation,chen2025conformalsam}, the teacher and student models respectively receive weak and strong augmentations of the same frame. However, the unsupervised training on unlabeled video data often results in limited viewpoint variation. Therefore, sequence-level augmentation is necessary to increase geometric diversity and prevent model collapse. 
Specifically, we first sample $24\!-\!36$ frames and shuffle the order before feeding them to the teacher. 
For the student branch, the same frames are randomly subsampled to $2\!-\!12$ images, shuffled again, and each image is rotated by $0^\circ$, $90^\circ$, $180^\circ$, or $270^\circ$ to avoid pose collapse.
We additionally apply color jittering, grayscale conversion, and Gaussian blur.

\subsection{Consistency-aware Training Objective}
\label{sec:loss}

3D reconstruction data generated by SfM pipelines or rendering engines inherently satisfy strong multi-view geometric consistency. 
To preserve this structural property during training, we introduce additional consistency objectives that encourage coherent geometric predictions. Our training objective consists of three components: a supervised loss following VGGT, a per-view consistency loss using teacher predictions as pseudo-labels, and a cross-view consistency loss enforcing multi-view geometric coherence. 
The overall training objective is defined as:
\begin{equation}
\mathcal{L}_{\mathrm{total}}
=
\mathcal{L}_{\mathrm{supervised}}
+
\lambda_{p}
\mathcal{L}_{\mathrm{per\text{-}view}}
+
\lambda_{c}\mathcal{L}_{\mathrm{cross\text{-}view}}.
\end{equation}

\mypar{Supervised Loss} 
The supervised loss is applied when training on labeled synthetic data, which is computed between the student predictions and the ground-truth annotations $\{g^{gt},D^{gt},P^{gt}\}$. Here we directly use the loss in VGGT~\cite{wang2025vggt}:

\begin{equation}
\mathcal{L}_{\mathrm{supervised}}
=
\mathcal{L}_{\mathrm{camera}}(g^{s}, g^{gt})
+
\mathcal{L}_{\mathrm{depth}}(D^{s}, D^{gt})
+
\mathcal{L}_{\mathrm{point}}(P^{s}, P^{gt}),
\end{equation}
where the depth loss includes regression, confidence, and gradient terms; the point loss consists of regression, confidence, and normal terms; and the camera loss is computed using Huber loss. Details can refer to VGGT~\cite{wang2025vggt}.

\mypar{Per-view Geometry Consistency Loss} We use this loss when learning from unlabeled real data.  
It shares the same formulation as the supervised loss by regarding the predictions of the teacher model, \emph{i.e.,} $\{g^{t},D^{t},P^{t}\}$, as pseudo ground truth. Notably, point maps are projected from depth and camera parameters for more stable geometric supervision~\cite{wang2025vggt}. 
The per-view loss is defined as:
\begin{equation}
\mathcal{L}_{\mathrm{per\text{-}view}}
=
\mathcal{L}_{\mathrm{camera}}(g^{s}, g^{t})
+
\mathcal{L}_{\mathrm{depth}}(D^{s}, D^{t})
+
\mathcal{L}_{\mathrm{point}}(P^{s}, P^{t}).
\end{equation}

\mypar{Cross-view Geometry Consistency Loss}
Underwater scattering and attenuation often lead to poor visibility and limited geometric information. To address this challenge, we introduce a cross-view geometry consistency loss that explicitly integrates geometric cues from other views to compensate for the information degradation in the current view.
Given $N$ teacher frames, we first obtain a coarse foreground mask $M^{\mathrm{fg}}_i$ for each frame $i$ using a simple two-cluster K-means on the depth map, keeping the closer cluster as foreground.
For every frame $i$, we backproject valid teacher pixels into 3D and reproject them into all other teacher views.
Define the projected pixel and the visibility indicator as:

\begin{equation}
\begin{aligned}
V_{i\rightarrow j}(x)
&=\mathbf{1}\!\left(
|D^t_j(u_{i\rightarrow j}(x)) - D^t_{i\rightarrow j}(x)| < \delta
\;\land\;
u_{i\rightarrow j}(x)\in\Omega_j
\right), \\
u_{i\rightarrow j}(x)
&=\pi_j\!\left(\pi^{-1}_i(x, D^t_i(x))\right).
\end{aligned}
\end{equation}
Here, $V_{i\!\rightarrow j}(x)$ indicates whether the pixel $x$ in frame $i$ remains visible and depth-consistent after being projected into frame~$j$. $\mathbf{1}(\cdot)$ returns $1$ if it is true.  $D^t_i$ denotes the teacher depth for view $i$. $\Omega_j$ is the image domain of view $j$. $u_{i\!\rightarrow j}(x)$ indicates the projection from view $i$ to view $j$. Wherein $\pi_j$ is the projection operator that projects the depth of view $j$ into the world coordinate system using the camera pose, and $\pi_i^{-1}$ is the back-projection operator. Then the statistic region of view $i$ is computed as the depth-consistent part of its foreground mask:

\begin{equation}
M^{\mathrm{static}}_i(x)
=
\mathbf{1}\!\left(
\sum_{j\neq i} V_{i\!\rightarrow j}(x)\ge k
\right)M^{\mathrm{fg}}_i(x).
\end{equation}

We use the conservative setting $k=N-2$ in our experiments, where $N$ denotes the number of teacher frames used to construct the mask.
This mask is not an additional supervision source; it only selects reliable pixels for applying cross-view supervision.
The foreground term suppresses distant background water regions, while the multi-view consistency term removes dynamic or geometrically unstable regions.
When the scene contains mostly open water, severe turbidity, or strong object motion, the mask naturally becomes sparse and therefore prevents the model from enforcing unreliable cross-view constraints.

During student training, we randomly select $n$ frames from the same sequence (in arbitrary order), forming an index set $\mathcal{S}$.  
For any pair $(i,j)\in\mathcal{S}$, depth from teacher frame $i$ is projected into view $j$:
\begin{equation}
D^{t}_{i\!\rightarrow j}(x)
=\pi_{j}\!\left(\pi^{-1}_{i}(x, D^{t}_i(x))\right),
\end{equation}
and compared with the student depth $D^s_j$ at frame $j$ under the static mask $M^{\mathrm{static}}_i$.
The geometry-consistent loss averages over all ordered pairs:
\begin{equation}
\mathcal{L}_{\mathrm{cross\text{-}view}}
=
\frac{1}{|\mathcal{S}|^2}
\sum_{i\in\mathcal{S}}
\sum_{\substack{j\in\mathcal{S},j\neq i}}
\mathcal{L}_{\mathrm{L1}}\!\left(
D^s_j,\,
D^{t}_{i\!\rightarrow j},\,
M^{\mathrm{static}}_j
\right).
\end{equation}

By now, we have introduced the main process of the proposed \textbf{Wat3R},  a cross-domain semi-supervised learning framework to generalize VGGT to the challenging underwater domain without requiring any underwater annotations.

\section{Experiments}

Next, we evaluate our \methodname{} on four underwater tasks: multi-view depth estimation (\cref{sec:mvsdepth}), point map estimation (\cref{sec:mvspoint}), camera pose estimation (\cref{sec:mvspose}), and monocular depth estimation (\cref{sec:singledepth}).
\subsection{Implementation Details}

We train the model on 4 NVIDIA RTX 4090 GPUs for 19,200 steps, beginning with a 1,000-step linear warm-up. The peak learning rates are set to $5 \times 10^{-6}$ for the ViT backbone and $5 \times 10^{-5}$ for the downstream head. To improve training efficiency and reduce GPU memory consumption, we employ gradient checkpointing and bfloat16 mixed-precision training. 
The ratio of unlabeled to labeled samples is set to 1:3. For the first 6,400 steps, training is conducted exclusively on labeled data. The unsupervised loss weight $\lambda_{u}$ is then gradually increased, reaching its peak value of 0.5 at step 12,800. During training, we randomly sample 2 to 12 images, with the longest side of each image resized to 518 pixels.

\begin{table}[t]
\centering
\caption{\textbf{Multi-view depth estimation.}
We report results under two evaluation protocols:
(1) Shuffle 10-view evaluation,
and (2) Full subsequence evaluation (100 frames, ordered).
The best and second of each category are masked as \textbf{Bold} and \underline{Underline}, respectively. 
{\setlength{\fboxsep}{0pt}\protect\colorbox{gray!15}{\strut Shaded rows}} denote two-stage pipelines, where input images are first enhanced by the corresponding underwater image enhancement method and then fed into VGGT model.
Values in {\dplustem{red}} indicate the percentage improvement over VGGT.}
\vspace{-2mm}
\scriptsize
\setlength{\tabcolsep}{1pt}

\begin{tabular}{@{}l cccc cccc}
\toprule
\multirow{3}{*}{\textbf{Methods}} &
\multicolumn{4}{c}{\textbf{Sea-thru~\cite{akkaynak2019sea}}} &
\multicolumn{4}{c}{\textbf{FLSea Stereo~\cite{randall2023flsea}}} \\
\cmidrule(lr){2-5} \cmidrule(lr){6-9}

& Rel$\downarrow$ & $\delta_1\uparrow$ & $\log_{10}\downarrow$
& RMSE$\downarrow$
& Rel$\downarrow$ & $\delta_1\uparrow$ & $\log_{10}\downarrow$
& RMSE$\downarrow$ \\

\midrule
\multicolumn{9}{c}{\textbf{Shuffle 10-view Evaluation (stride$=$10)}} \\
\midrule    

WaterSplatting~\cite{li2025watersplatting}~\tiny{3DV\textquotesingle25} & - & - & - & - & 0.427

 & 0.415

 & 0.157
 & 1.476
\\

Fast3r~\cite{yang2025fast3r}~\tiny{CVPR\textquotesingle25} &
0.277 & 0.713 & 0.083 & 0.631 &
0.290 & 0.529 & 0.141 & 1.355 \\

MapAnything~\cite{keetha2025mapanything}~\tiny{3DV\textquotesingle26} &
0.216 & 0.800 & 0.060 & 0.454 &
0.146 & 0.841 & 0.061 & 0.798 \\

$\pi3$~\cite{wang2025pi}~\tiny{ICLR\textquotesingle26} &
\underline{0.185} & \underline{0.909} & \underline{0.044} &
0.358 &
0.139 & 0.856 & \underline{0.053} & 0.837 \\

DA3~\cite{lin2025depth}~\tiny{ICLR\textquotesingle26} &
0.187 & 0.892 & 0.046 & \underline{0.333} &
0.141 & 0.851 & 0.056 & 0.872 \\

VGGT~\cite{wang2025vggt}~\tiny{CVPR\textquotesingle25} &
0.190 & 0.891 & 0.047 & 0.380 &
0.137 & 0.849 & 0.059 & \underline{0.760} \\
\rowcolor{gray!15}~+ Semi-UIR~\cite{huang2023contrastive}~\tiny{CVPR\textquotesingle23}
& 0.201& 0.846& 0.052& 0.387 & 
\underline{0.136} & \underline{0.861} & 0.056 & 0.784

\\
\rowcolor{gray!15}~+ PSPL~\cite{liu2025toward}~\tiny{TIP\textquotesingle25}
& 0.209& 0.836& 0.055& 0.419 & 
0.160 & 0.817 & 0.063 & 0.911

\\

\multirow{2}{*}{\textbf{\methodname}} &
\textbf{0.167} & \textbf{0.946} & \textbf{0.038} & \textbf{0.290} &
\textbf{0.119} & \textbf{0.885} & \textbf{0.048} & \textbf{0.720} \\[-5pt]
& \dplus{+12.1\%} & \dplus{+6.2\%} & \dplus{+19.1\%} & \dplus{+23.7\%} &
\dplus{+13.1\%} & \dplus{+4.2\%} & \dplus{+18.6\%} & \dplus{+5.3\%} \\

\midrule
\multicolumn{9}{c}{\textbf{Full Subsequence Evaluation (100 frames, ordered)}} \\
\midrule

Fast3r~\cite{yang2025fast3r}~\tiny{CVPR\textquotesingle25} &
0.274 & 0.711 & 0.081 & 0.622 &
0.299 & 0.521 & 0.146 & 1.363 \\

MapAnything~\cite{keetha2025mapanything}~\tiny{3DV\textquotesingle26} &
0.227 & 0.788 & 0.063 & 0.488 &
0.147 & 0.841 & 0.062 & 0.795 \\

$\pi3$~\cite{wang2025pi}~\tiny{ICLR\textquotesingle26} &
0.187 & \underline{0.923} & 0.041 &
0.358 &
0.139 & \underline{0.856} & \underline{0.053} &
0.836 \\

DA3~\cite{lin2025depth}~\tiny{ICLR\textquotesingle26} &
\underline{0.183} & 0.922 & \underline{0.040} & \underline{0.311} &
0.139 & 0.855 & 0.055 & 0.861 \\

VGGT~\cite{wang2025vggt}~\tiny{CVPR\textquotesingle25} &
0.193 & 0.900 & 0.045 & 0.373 &
\underline{0.136} & 0.851 & 0.058 & \underline{0.759} \\

\multirow{2}{*}{\textbf{\methodname}} &
\textbf{0.170} & \textbf{0.953} & \textbf{0.036} & \textbf{0.294} &
\textbf{0.120} & \textbf{0.884} & \textbf{0.048} & \textbf{0.715} \\[-5pt]
& \dplus{+11.9\%} & \dplus{+5.9\%} & \dplus{+20.0\%} & \dplus{+21.2\%} &
\dplus{+11.8\%} & \dplus{+3.9\%} & \dplus{+17.2\%} & \dplus{+5.8\%} \\

\bottomrule
\end{tabular}

\vspace{-0.15in}
\label{tab:multi_result}
\end{table}

\mypar{Datasets} To ensure applicability to underwater vision scenarios, we conduct experiments on five public datasets together with our own constructed \datasetname{}:
\begin{itemize}
\item \textbf{FLSea-VI~\cite{randall2023flsea}}
contains forward-looking monocular underwater sequences captured in shallow Mediterranean and Red Sea environments. 
We evaluate on the Coral Table Loop and U-Canyon sequences.

\item \textbf{FLSea-Stereo~\cite{randall2023flsea}}
comprises 7,341 synchronized stereo image pairs from 5 shallow-water scenes, including canyon and flat sandy seabed environments, collected from the FLSea-VI dataset in different dives and scenes.

\item \textbf{SQUID~\cite{berman2020underwater}}
provides a quantitative underwater stereo dataset consisting of 57 stereo image pairs with stereo-derived ground-truth depth, collected at different seasons, depths, and water types in natural marine environments.
\item \textbf{Sea-thru~\cite{akkaynak2019sea}}
is an underwater RGBD dataset composed of 1,205 images from 5 natural scenes, acquired in two optically different water bodies under natural illumination, spanning depths, water types, and scene structures.

\item \textbf{SeaThru-NeRF~\cite{levy2023seathru}}
consists of 5 real underwater scenes annotated with camera poses only. This dataset is originally used for NeRF evaluation.

\item \textbf{\datasetname} consists of 42 real underwater scenes, where 20 are from UVEB~\cite{xie2024uveb}, and 22 scenes are collected alongside our training data (but not used for training). Camera poses and depths are reconstructed using COLMAP~\cite{schonberger2016structure}, wherein the matcher is changed to the advanced MINIMA~\cite{ren2025minima}, which is further fine-tuned with our synthetic underwater data. 
The final results are manually checked to make sure the correctness of annotations, where only the scenes with stable registration and no obvious collapse or severe outliers are preserved. \textit{See supplement for more details.}

\end{itemize}

\begin{figure*}[t]
  \centering
  \includegraphics[width=0.95\linewidth]{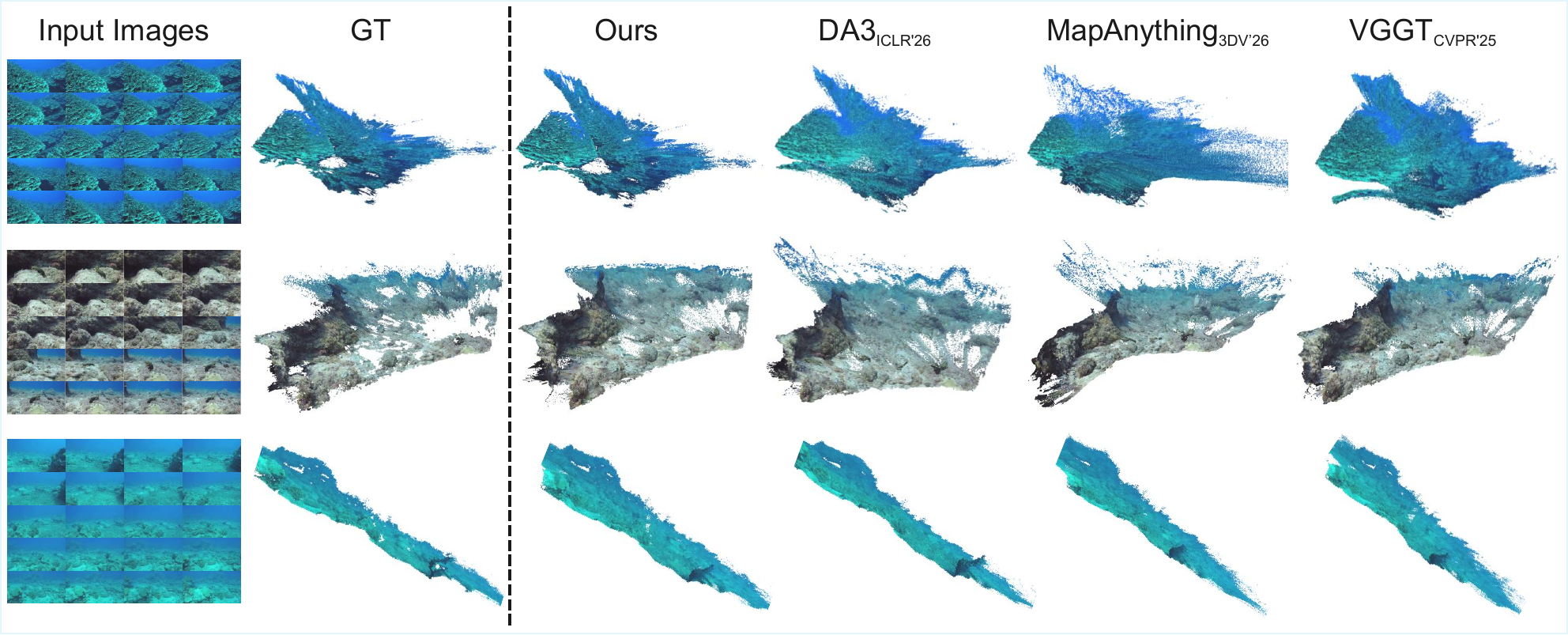}
    \vspace{-1mm}
  \caption{\textbf{3D reconstruction results on our constructed \datasetname{}}. DA3~\cite{lin2025depth}, MapAnything~\cite{keetha2025mapanything},  VGGT~\cite{wang2025vggt} and our \methodname{} are used for comparison. \methodname{} obtains more complete and physically consistent reconstructions with reduced artifacts.}
\label{fig:point_result}  
\end{figure*}

\begin{figure*}[t]
  \centering
  \includegraphics[width=0.95\linewidth]{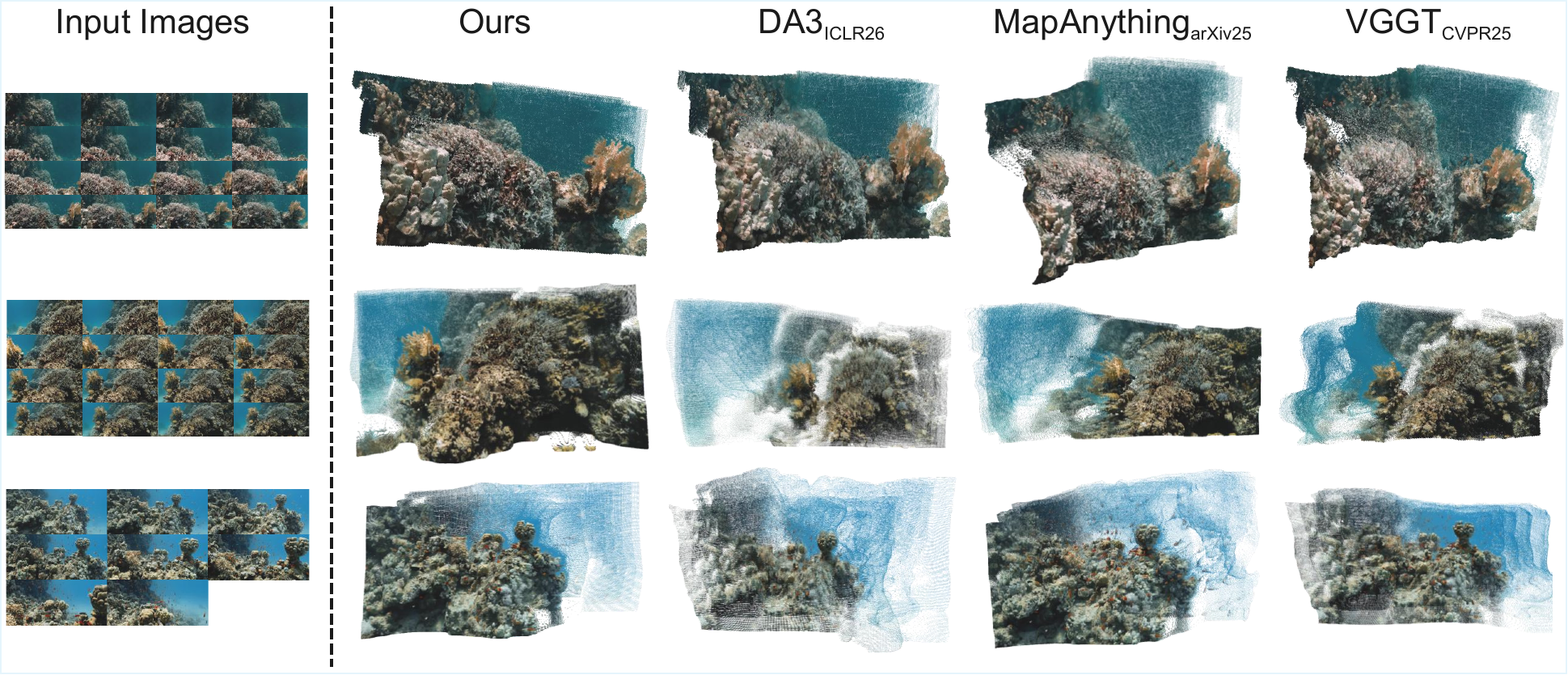}
    \vspace{-1mm}
  \caption{\textbf{Qualitative comparison on in-the-wild images} with \methodname{}, DA3~\cite{lin2025depth}, MapAnything~\cite{keetha2025mapanything}, and VGGT~\cite{wang2025vggt}. For a fair comparison, all methods take the same input. No post-processing is applied, allowing full point clouds to be visualized.}
\label{fig:wild_point_result}  \vspace{-0.15in}
\end{figure*}

\subsection{Multi-view Depth Estimation}
\label{sec:mvsdepth}

We first evaluate the depth estimation from multi-view input. Two video datasets with depth annotations are used, \emph{i.e.,} Sea-thru~\cite{akkaynak2019sea} and FLSea-Stereo~\cite{randall2023flsea}. 
We compare our Wat3R with state-of-the-art models trained on large-scale annotated data. Besides, the underwater Splatting method~\cite{li2025watersplatting} is also evaluated. To verify whether the underwater image enhancement (UIE) method is effective for 3D geometry estimation, we also take two enhancement methods into a two-stage pipeline. It first enhances underwater images to make them closer to the on-land domain, then feeds them into a 3D model. Following~\cite{wang2024dust3r,wang2025vggt,yang2025fast3r}, long sequences are divided into multiple sub-scenes, each containing at most 100 images. And we uniformly sample 10 images from each sequence and randomly shuffle their order for testing. In addition, we follow the video-based depth evaluation protocol of~\cite{chen2025video} to evaluate full subsequences. We evaluate metric depth by aligning predicted depth maps to the ground-truth scale using a least-squares estimation of scale and shift. Performance is measured using standard depth metrics, and results are reported in~\cref{tab:multi_result}.
We report absolute relative error (Rel), absolute error in log-scale ($\log_{10}$), root mean squared error (RMSE), and the percentage of inlier pixels $\delta_i$ under thresholds of $1.25^i$ to evaluate overall depth accuracy. The frame order of input video is randomly shuffled during training, so the model does not rely on temporal continuity or video-specific supervision.

\cref{tab:multi_result} reveals that our model achieves clear and consistent improvements compared to the original VGGT~\cite{wang2025vggt}. 
Specifically, WaterSplatting~\cite{li2025watersplatting} is a Splatting method designed for underwater scenes. But it relies on initialized point clouds, producing large errors on FLSea-Stereo and even failing on the challenging Sea-thru dataset. 
$\pi3$~\cite{wang2025pi} and DA3~\cite{lin2025depth} obtain competitive results due to their improved 3D representations, and their large training set contains some underwater scenes.
The shaded rows show that UIE almost brings no improvement for multi-view depth estimation. The enhancement process may easily introduce information loss or view-inconsistent changes, failing to recover new structural cues.
In contrast, \methodname{} learns underwater-aware geometry directly during training, leading to more reliable 3D perception in challenging underwater scenes.
  These results highlight our method's ability to achieve superior accuracy and robustness, particularly in challenging underwater multi-view scenarios.  These high-fidelity results are visually demonstrated in~\cref{fig:point_result} and \cref{fig:wild_point_result}.

\begin{table}[t]
\centering
\caption{\textbf{Performance comparison on the \datasetname{} dataset.} The best and second of each category are masked as \textbf{Bold} and \underline{Underline}, respectively. Values in {\dplustem{red}} indicate the percentage improvement over VGGT\cite{wang2025vggt}.}
\vspace{-2mm}
\scriptsize
\setlength{\tabcolsep}{1.8mm}
\begin{tabular}{@{}lcccccc}
\toprule
\multirow{3}{*}{\textbf{Methods}} & 
\multicolumn{2}{c}{\textbf{Accuracy} $\downarrow$} &
\multicolumn{2}{c}{\textbf{Completeness} $\downarrow$} & 
\multicolumn{2}{c}{\textbf{Overall} $\downarrow$} \\
\cmidrule(lr){2-3}
\cmidrule(lr){4-5}
\cmidrule(lr){6-7}
& Mean & Median
& Mean & Median
& Mean & Median \\
\midrule

Fast3r~\cite{yang2025fast3r}~\tiny{CVPR\textquotesingle25}
& 1.216 & 0.531 & 2.444 & 0.784 & 1.830 & 0.658 \\

MapAnything~\cite{keetha2025mapanything}~\tiny{3DV\textquotesingle26}
& 0.643 & 0.284 & 0.666 & 0.277 & 0.655 & 0.281 \\

$\pi3$~\cite{wang2025pi}~\tiny{ICLR\textquotesingle26}
& 0.491 & 0.168 & 0.413 & 0.184 & 0.452 & 0.176 \\

DA3~\cite{lin2025depth}~\tiny{ICLR\textquotesingle26}
& 0.679 & 0.187 & 0.528 & \underline{0.160} & 0.604 & 0.174 \\

\midrule

VGGT~\cite{wang2025vggt}~\tiny{CVPR\textquotesingle25}
& 0.486 & 0.193 & 0.762 & 0.191 & 0.624 & 0.192 \\

\multirow{2}{*}{\textbf{\methodname{}~(Point)}} &
\textbf{0.444} & \textbf{0.148} & \underline{0.409} & 0.165 & \underline{0.427} & \underline{0.157} \\[-5pt]
& \dplus{+8.6\%} & \dplus{+23.3\%} & \dplus{+46.3\%} & \dplus{+13.6\%} & \dplus{+31.6\%} & \dplus{+18.2\%} \\

\multirow{2}{*}{\textbf{\methodname{}~(Depth+Cam)}} &
\underline{0.446} & \underline{0.162} & \textbf{0.366} & \textbf{0.143} & \textbf{0.406} & \textbf{0.153} \\[-5pt]
& \dplus{+8.2\%} & \dplus{+16.1\%} & \dplus{+52.0\%} & \dplus{+25.1\%} & \dplus{+34.9\%} & \dplus{+20.3\%} \\

\bottomrule
\end{tabular}

\label{tab:uveb}
\end{table}

\subsection{Point Map Estimation}
\label{sec:mvspoint}

We further evaluate the quality of reconstructed multi-view point maps using our underwater dataset \datasetname{}. For each sequence, 20 images are uniformly sampled for evaluation. Following~\cite{yang2025fast3r,wang2025pi,wang2025vggt}, we first perform a coarse Sim(3) alignment using the Umeyama algorithm~\cite{umeyama1991least}, followed by refinement with the Iterative Closest Point (ICP) algorithm. We report Accuracy, Completeness, and their Overall average as metrics in~\cref{tab:uveb}. 
The direct point-map output and the reconstruction derived from predicted depth and cameras both achieve strong performance.
Their consistent results indicate that adaptation preserves geometric agreement among the point, depth, and camera heads.
This is enabled by our consistency-aware objectives, which couple predictions across heads and views even without ground-truth
annotations for real underwater videos.

\subsection{Camera Pose Estimation}
\label{sec:mvspose}

We now evaluate camera pose estimation on the SeaThru-NeRF~\cite{levy2023seathru}.
Following prior work~\cite{wang2025vggt,wang2024dust3r}, performance is measured using AUC under different angular thresholds, where AUC is calculated from the minimum of Relative Translation Accuracy (RTA) and Relative Rotation Accuracy (RRA). 

As shown in~\cref{tab:pose_result}, DA3~\cite{lin2025depth} achieves the best overall performance across all thresholds. This is mainly due to its depth-ray representation, which provides a minimal yet sufficient formulation for jointly modeling scene geometry and camera motion. Geometry-driven pose inference exhibits natural robustness in underwater environments.
Our cross-view consistency learning enables the model to extract complementary geometric cues across views, reducing its
sensitivity to water-induced degradation and yielding more stable pose estimates.

\begin{table}[t]
\centering
\caption{\textbf{Pose estimation AUC at different thresholds on the SeaThru-NeRF~\cite{levy2023seathru}.} The best and second of each category are masked as \textbf{Bold} and \underline{Underline}, respectively. Values in {\dplustem{red}} indicate the percentage improvement over VGGT\cite{wang2025vggt}.}
\scriptsize
\vspace{-1mm}

\setlength{\tabcolsep}{16.05pt}

\begin{tabular}{@{}lccc}
\toprule
\textbf{Methods} & \textbf{AUC@5$\degree$} & \textbf{AUC@15$\degree$} & \textbf{AUC@30$\degree$} \\
\midrule
Fast3r~\cite{yang2025fast3r}~\tiny{CVPR\textquotesingle25} & 0.040 & 0.239 & 0.498 \\

$\pi3$~\cite{wang2025pi}~\tiny{ICLR\textquotesingle26} & 0.216 & 0.635 & 0.809 \\
DA3~\cite{lin2025depth}~\tiny{ICLR\textquotesingle26} & \textbf{0.731} & \textbf{0.901} & \textbf{0.950} \\
MapAnything~\cite{keetha2025mapanything}~\tiny{3DV\textquotesingle26} & 0.074 & 0.458 & 0.707 \\
\midrule

VGGT~\cite{wang2025vggt}~\tiny{CVPR\textquotesingle25} & 0.392 & 0.707 & 0.843 \\

\textbf{\methodname}
& \underline{0.540}\dplustem{+37.8\%}
& \underline{0.820}\dplustem{+16.0\%}
& \underline{0.906}\dplustem{+7.5\%} \\
\bottomrule
\end{tabular}
\vspace{-0.1in}
\label{tab:pose_result}
\end{table}

\begin{table}[t]
\centering
\caption{\textbf{Monocular depth estimation.} The best and second of each category are masked as \textbf{Bold} and \underline{Underline}, respectively. The top three methods (Udepth~\cite{yu2022udepth}, UW-Depth~\cite{ebner2024metrically}, WaterMono~\cite{ding2025watermono}) are underwater-specific models. WaterMono~\cite{ding2025watermono} is trained on FLSea VI~\cite{randall2023flsea}; its results are shown in {\color{gray}gray}. Values in {\dplustem{red}} indicate the percentage improvement over VGGT\cite{wang2025vggt}.}
\vspace{-0.5mm}
\scriptsize
\setlength{\tabcolsep}{0.2pt}
\begin{tabular}{@{}l cc cc cc cc}
\toprule
\multirow{3}{*}{\vspace{+0.5em}\textbf{Methods}} & 
\multicolumn{2}{c}{\textbf{FLSea VI~\cite{randall2023flsea}}}  &
\multicolumn{2}{c}{\textbf{FLSea Stereo~\cite{randall2023flsea}}}  &
\multicolumn{2}{c}{\textbf{SQUID~\cite{berman2020underwater}}} & 
\multicolumn{2}{c}{\textbf{Sea-thru~\cite{akkaynak2019sea}}} 
\\
\cmidrule(lr){2-3}
\cmidrule(lr){4-5}
\cmidrule(lr){6-7}
\cmidrule(lr){8-9}

& 
Rel$\downarrow$ & $\delta_{1}\uparrow$ & 
Rel$\downarrow$ & $\delta_{1}\uparrow$ & 
Rel$\downarrow$ & $\delta_{1}\uparrow$ & 
Rel$\downarrow$ & $\delta_{1}\uparrow$ 
\\
\midrule
Udepth~\cite{yu2022udepth}~\tiny{ICRA\textquotesingle23}
& 0.212 & 0.683 & 0.279 & 0.550 & 0.312 & 0.547 & 0.166	 & 0.832
\\
UW-Depth~\cite{ebner2024metrically}~\tiny{ICRA\textquotesingle24}
& 0.330 & 0.447 & 0.400 & 0.427 & 0.491 & 0.343 & 0.184 & 0.792

\\
WaterMono~\cite{ding2025watermono}~\tiny{TIM\textquotesingle25}
& {\color{gray}0.074} & {\color{gray}0.949} & 0.206 & 0.684 & 0.261 & 0.544 & 0.145 & 0.829
\\

\midrule
DAv2~\cite{yang2024depth}~\tiny{NeurIPS\textquotesingle24}
& \underline{0.069} & \underline{0.958} & \underline{0.134} & \underline{0.849} & \textbf{0.099} & \textbf{0.900} & \textbf{0.089} & 0.940
\\
Fast3r~\cite{yang2025fast3r}~\tiny{CVPR\textquotesingle25}
& 0.198 & 0.720 & 0.213 & 0.679 & 0.368 & 0.522 & 0.125 & 0.911
\\

MapAnything~\cite{keetha2025mapanything}~\tiny{3DV\textquotesingle26}
& 0.086 & 0.951 & 0.146 & 0.837 & \underline{0.104} & \underline{0.898} & 0.104 & \underline{0.960} 

\\
$\pi3$~\cite{wang2025pi}~\tiny{ICLR\textquotesingle26}
& 0.081 & 0.944 & 0.148 & 0.831 & 0.234 & 0.640 & 0.113 & 0.949
\\
DA3~\cite{lin2025depth}~\tiny{ICLR\textquotesingle26}
& 0.090 & 0.936 & 0.154 & 0.824 & 0.151 & 0.802 & 0.111 & 0.950
\\

\midrule
VGGT~\cite{wang2025vggt}~\tiny{CVPR\textquotesingle25} 
& 0.107 & 0.888 & 0.159 & 0.809 & 0.194 & 0.703 & 0.113 & 0.942
\\
\multirow{2}{*}{\textbf{\methodname}} &
\textbf{0.061} & \textbf{0.971} & \textbf{0.120} & \textbf{0.886} &
0.107 & 0.893 & \underline{0.090} & \textbf{0.976} \\[-5pt]
& \dplus{+43.0\%} & \dplus{+9.3\%} & \dplus{+24.5\%} & \dplus{+9.5\%} &
\dplus{+44.8\%} & \dplus{+27.0\%} & \dplus{+20.4\%} & \dplus{+3.6\%} \\
\bottomrule
\end{tabular}
\label{tab:monocular_result2}
\end{table}

\begin{figure*}[t]
  \centering
  \includegraphics[width=0.95\linewidth]{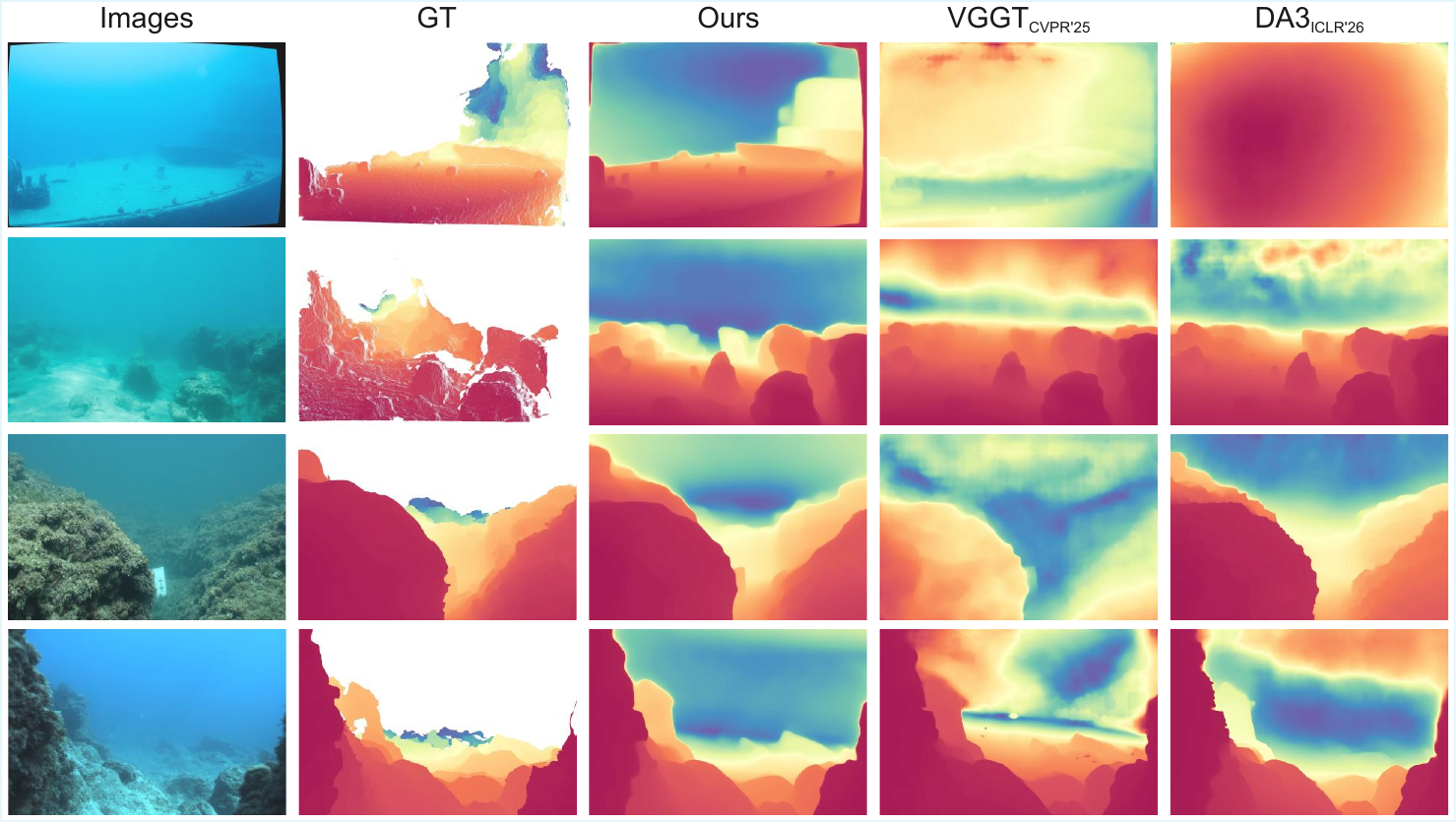}
    \vspace{-2mm}
    \caption{\textbf{Monocular depth estimation using \methodname{}, DA3~\cite{lin2025depth}, and VGGT~\cite{wang2025vggt}}. The first two rows are sampled from the SQUID~\cite{berman2020underwater} dataset, the third row from FLSea-Stereo~\cite{randall2023flsea}, and the fourth row from FLSea-VI~\cite{randall2023flsea}.}
\label{fig:depth_result}  
\end{figure*} 
\vspace{-2mm}

\subsection{Monocular Depth Estimation}
\label{sec:singledepth}

We evaluate monocular depth estimation to assess the model's understanding of underwater degradation.
In this setting, no complementary cues from other views are available.
Following~\cite{lin2025unified,zhang2024atlantis}, we evaluate performance on four datasets: FLSea-VI~\cite{randall2023flsea}, FLSea-Stereo~\cite{randall2023flsea}, SQUID~\cite{berman2020underwater} and Sea-thru~\cite{akkaynak2019sea}, using Absolute Relative Error (Rel) and $\delta_1$ as metrics. The results are summarized in~\cref{tab:monocular_result2}.
Although our model is not trained with single-image supervision, it achieves the best performance across four datasets.
The cross-view supervision helps separate underwater degradation from scene geometry.
The learned geometric prior then transfers to single-view input.
~\cref{fig:depth_result} presents qualitative monocular results. 

\begin{table}[t]
\centering
\scriptsize
\caption{\textbf{Ablation studies.} The multi-view depth estimation is evaluated. The best and second of each category are masked as \textbf{Bold} and \underline{Underline}, respectively.}
\vspace{-2mm}
\setlength{\tabcolsep}{4.5pt}
\begin{tabular}{lccccc|cccc}
\toprule

& \multirow{2}{*}[-0.35em]{\makecell{Syn\\Water}} &
\multirow{2}{*}[-0.35em]{\makecell{Real\\Video}} &
\multirow{2}{*}[-0.35em]{\makecell{Strong\\Aug.}} &
\multirow{2}{*}[-0.35em]{\makecell{$\mathcal{L}_{\mathrm{cross\text{-}view}}$}} &
\multirow{2}{*}[-0.35em]{\makecell{$M^{\mathrm{static}}$}} &

\multicolumn{2}{c}{\textbf{Sea-thru}} &
\multicolumn{2}{c}{\textbf{FLSea Stereo}} \\

\cmidrule(lr){7-8} \cmidrule(lr){9-10}

& & & & &
& Rel$\downarrow$ & $\delta_{1}\uparrow$
& Rel$\downarrow$ & $\delta_{1}\uparrow$ \\

\midrule

\textbf{VGGT}
&  &  &  &  &  & 0.190 & 0.891 & 0.137 & 0.849 \\

& $\checkmark$ &  &  &  &  & 0.173 & 0.920 & 0.135 & 0.860 \\

& $\checkmark$ & $\checkmark$ &  &  & 
& 0.181 & 0.906 & 0.165 & 0.793 \\

& $\checkmark$ & $\checkmark$ & $\checkmark$ &  & 
& \underline{0.172} & 0.936 & \underline{0.126} & \underline{0.871} \\

& $\checkmark$ & $\checkmark$ & $\checkmark$ & $\checkmark$ & 
& \textbf{0.167} & \textbf{0.949} & \underline{0.126} & 0.869 \\

& $\checkmark$ &  & $\checkmark$ & $\checkmark$ & $\checkmark$
& 0.174 & 0.929 & 0.130 & \underline{0.871} \\

\textbf{\methodname{}}
& $\checkmark$ & $\checkmark$ & $\checkmark$ & $\checkmark$ & $\checkmark$
& \textbf{0.167} & \underline{0.946} & \textbf{0.119} & \textbf{0.885} \\

\bottomrule
\end{tabular} \vspace{-0.2in}
\label{tab:ablation333333}
\end{table}

\subsection{Ablation Studies}
\label{sec:ablation}

\cref{tab:ablation333333} presents an ablation study to analyze the contribution of each component in our framework. Starting from the pretrained VGGT model as the baseline, we progressively introduce key elements of our method, including real underwater video data, sequence-level augmentation, the proposed cross-view consistency loss $\mathcal{L}_{\mathrm{cross\text{-}view}}$, and the static mask $M^{\mathrm{static}}$. For reference, we include a supervised variant trained only on synthetic underwater data to highlight the effect of unlabeled real underwater videos and the proposed consistency constraints.

The results show three clear trends.
(1) Training with synthetic underwater rendering already improves performance over the VGGT baseline by partially bridging the domain gap between terrestrial and underwater imagery. 
(2) Incorporating real underwater video data together with strong augmentation consistently improves performance, demonstrating that large-scale unlabeled videos provide effective supervision signals for adapting geometry models to underwater scenes. 
(3) The proposed cross-view consistency loss and static mask further improve performance by enforcing multi-view geometric coherence and suppressing unstable regions caused by scattering or dynamic content.
Note that our masking strategy primarily deals with severe underwater degradation and dynamic objects. For simpler and static scenes like Sea-thru, the gains are not significant.

\begin{figure}[t]
  \centering
  \includegraphics[width=0.95\linewidth]{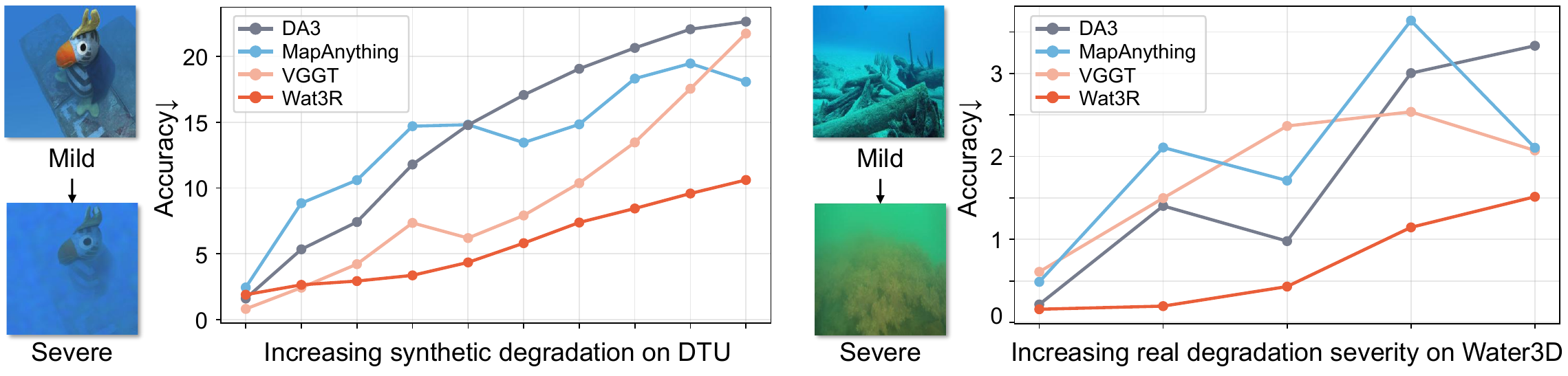}
  \vspace{-2mm}
  \caption{\textbf{Robustness and failure case analysis.} Left: synthetic underwater degradation with increasing visual corruption.
  Right: real underwater scenes grouped by difficulty.
  \methodname{} shows more robust to moderate degradation.}
  \label{figs:robust}
\end{figure}
\vspace{-0.15in}

\subsection{Robustness Analysis and Failure Cases}
\label{sec:robust_exp}
To evaluate the robustness of the model under extreme conditions, we report results under progressively stronger underwater degradation on both synthetic and real scenes in \cref{figs:robust}.
On synthetically degraded DTU~\cite{dtudataset} dataset, the point-map error of all methods increases as attenuation and scattering intensify, whereas \methodname{}
degrades more gradually and maintains a clear margin over the competing methods.
This advantage also extends to real \datasetname{} scenes.
Severe visibility loss reduces the stable visual evidence available for geometry recovery.
Nevertheless, \methodname{} deteriorates more slowly. 
Its robustness stems from training on diverse synthetic and real underwater conditions, together with consistency objectives
that encourage geometry to remain stable under appearance changes.

\subsection{Discussion on Possible Limitations}
Although \methodname{} is designed for underwater geometry reconstruction without requiring underwater annotations, several limitations remain.
Handling highly dynamic elements, such as moving marine life or divers, may require additional modeling of temporal dynamics and motion cues. In open-water, highly turbid or very deep scenes, reliable matched structures can become sparse. Our conservative static mask then suppresses unreliable regions, but this also reduces the cross-view learning signal and limits performance in such extreme cases.

\section{Conclusion}

In this paper, we introduced \methodname{}, a first semi-supervised VGGT framework for underwater geometry learning. \methodname{} learns robust geometry representations merely on unlabeled real underwater video, without the need for any annotated underwater data. And a cross-view consistency loss is designed to address underwater degradation for better training. Besides, we construct {\datasetname{}}, the first comprehensive evaluation benchmark containing diverse underwater conditions. Experimental results demonstrate that \methodname{} significantly outperforms the state-of-the-art in underwater multi-view geometry estimation.


\section*{Acknowledgements}
This work was supported by the National Natural Science Foundation of China (Grant U2341227, 62406117, and U25B2078).

%
%
\bibliographystyle{splncs04}
\bibliography{main}

\clearpage
\appendix
\setcounter{figure}{0}
\setcounter{table}{0}
\renewcommand\thefigure{A\arabic{figure}}    
\renewcommand\thetable{A\arabic{table}} 
\appendix
\makeatletter
\@ifundefined{theHsection}{}{%
  \renewcommand{\theHsection}{appendix.\Alph{section}}%
  \renewcommand{\theHsubsection}{appendix.\Alph{section}.\arabic{subsection}}%
  \renewcommand{\theHfigure}{appendix.\Alph{section}.\arabic{figure}}%
  \renewcommand{\theHtable}{appendix.\Alph{section}.\arabic{table}}%
  \renewcommand{\theHequation}{appendix.\Alph{section}.\arabic{equation}}%
}
\makeatother

\begin{center}
    \textbf{\Large Supplementary Material for ``Wat3R: Underwater\\ 3D Geometry Learning without Annotations''}
    \\ [0cm]
\end{center}

\crefname{section}{Appendix}{Appendices}
\Crefname{section}{Appendix}{Appendices}
In this appendix, we provide additional analyses and implementation details of our method:

\begin{itemize}
    \item analysis of the cross-view supervision mechanism in~\cref{appendix:loss};
    
    \item construction pipeline of the \datasetname{} dataset in~\cref{appendix:dataset};
    
    \item visualization of the proposed physics-based synthetic underwater rendering process in~\cref{appendix:render};
    
    \item further analysis of \methodname{}, including condition-wise results, comparisons with underwater-specific 3DGS and classical SfM pipelines, and additional ablations in~\cref{appendix:understanding};
    
    \item impact of underwater image enhancement on 3D reconstruction in~\cref{appendix:uie};
    
    \item additional qualitative results for monocular depth estimation and multi-view 3D reconstruction in~\cref{appendix:vis}.
\end{itemize}

\section{Analysis of Cross-view Supervision}
\label{appendix:loss}

\begin{figure*}[t]
  \centering
  \includegraphics[width=0.95\linewidth]{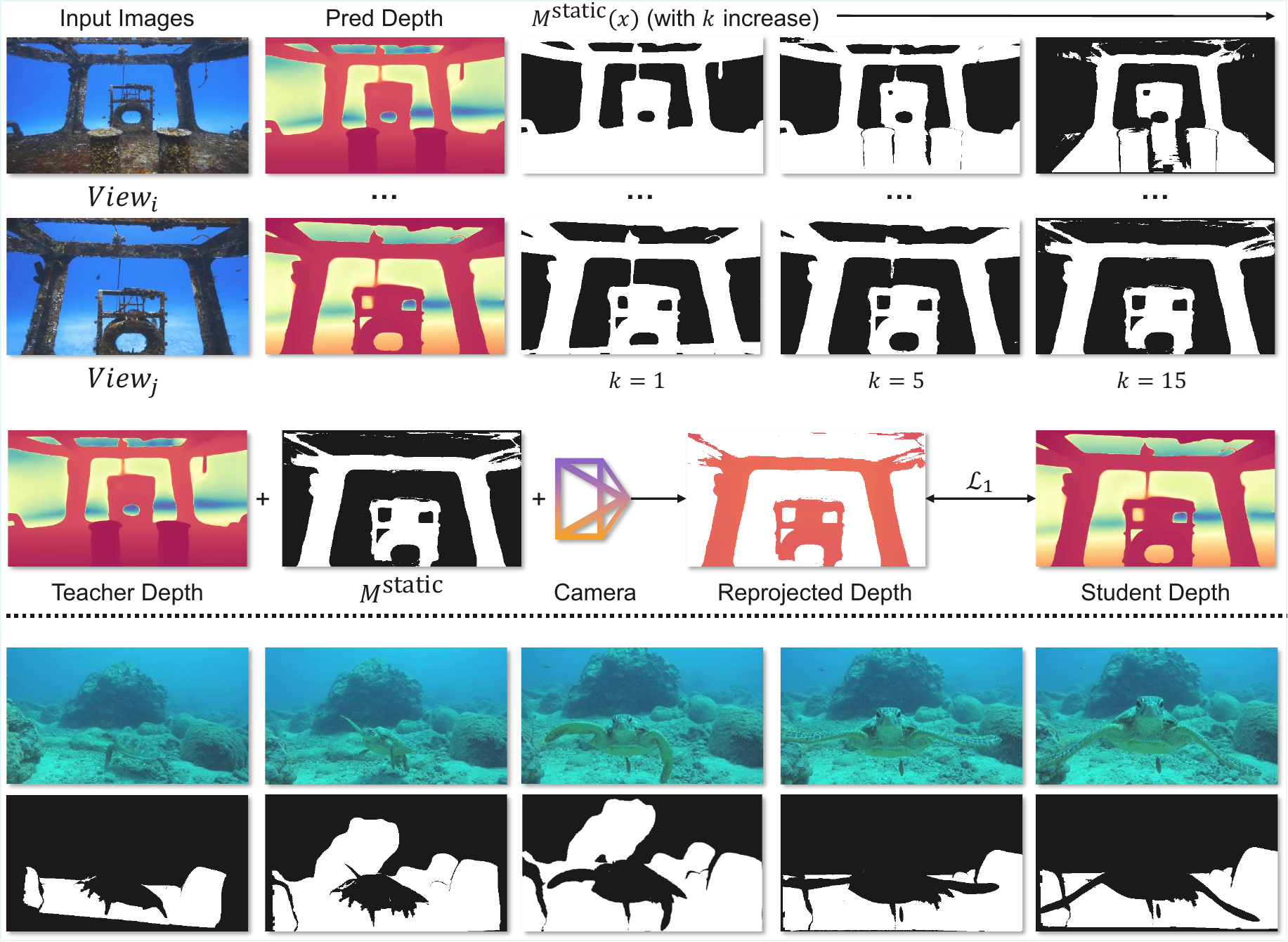}
\caption{\textbf{Cross-view geometric supervision with static region filtering.}
Teacher depth is reprojected from view $i$ to view $j$ to supervise student predictions. A multi-view static mask $M^{\mathrm{static}}$ is constructed using a consistency threshold $k$ to filter dynamic or inconsistent regions. The bottom row shows an example of a dynamic scene where moving objects are filtered by the static mask.}
\label{figs:cross_view_loss}  
\end{figure*}

To better understand cross-view supervision, we illustrate its pipeline and the effect of the static mask in dynamic scenes  in~\cref{figs:cross_view_loss}. We compare different supervision sources for the cross-view alignment in~\cref{tab:sup_cross_view_loss} 
The first row removes the cross-view loss entirely, serving as the baseline. 
The second row applies self-supervision by constructing the cross-view constraint using the student predictions themselves. 
The third row uses the teacher predictions as the supervision target.
The results show that the introduction of cross-view supervision improves performance compared to the baseline without loss. 
However, using student predictions as the target leads to limited improvement due to the instability of the predictions during training. 
This demonstrates that the EMA teacher produces more reliable geometric targets for enforcing cross-view consistency.

\section{\datasetname{} Dataset Details Construction}
\label{appendix:dataset}
In this section, we introduce the construction pipeline of \datasetname{}, from raw video streams to reconstructed point clouds. First, images are extracted from the raw videos at 10 frames per second. We use the hloc toolbox~\cite{sarlin2019coarse} to build the COLMAP~\cite{schonberger2016structure} reconstruction pipeline. To increase the accuracy of hloc, we adopt the MINIMA$_{\text{LightGlue}}$~\cite{lindenberger2023lightglue,ren2025minima} model finetuned with our synthetic underwater data. The resulting correspondences are passed to COLMAP to perform structure-from-motion reconstruction, producing camera intrinsics, extrinsics, and sparse point clouds. For dense reconstruction, we employ COLMAP's PatchMatch Stereo~\cite{bleyer2011patchmatch} algorithm. The parameters are adjusted to better handle weak-texture regions in underwater environments. Multi-view geometric consistency is enforced during depth estimation, and unreliable depth predictions are filtered based on photometric and geometric constraints.

Due to the presence of suspended particles in underwater scenes, raw depth maps often contain significant outliers. We project the depth maps into 3D space and use a KD-tree for nearest-neighbor analysis to remove outliers with low local density. The filtered points are then re-projected to refine the depth maps.

\begin{table*}[h]
\centering
\footnotesize
\setlength{\tabcolsep}{1.8mm}
\caption{\textbf{Effect of Different Supervision Sources for the Cross-view Loss.} We report Absolute Relative Error (Rel$\downarrow$) and $\delta_{1}\uparrow$ for both multi-view depth and monocular depth estimation on FLSea Stereo~\cite{randall2023flsea}.}
\begin{tabular}{@{}l cc cc}

\toprule
 \multirow{3}{*}{\vspace{+0.5em}\textbf{Cross-view supervision}}
 & \multicolumn{2}{c}{\textbf{Multi-view Depth}} 
 & \multicolumn{2}{c}{\textbf{Monocular Depth}} 
 \\
\cmidrule(lr){2-3} \cmidrule(lr){4-5} 
& Rel$\downarrow$ & $\delta_{1}\uparrow$
& Rel$\downarrow$ & $\delta_{1}\uparrow$
\\
\midrule

None & 0.126 & 0.871 & 0.127 & 0.871
\\
Student prediction & 0.122 & 0.881 & 0.123 & 0.880

\\
Teacher prediction & \textbf{0.119} & \textbf{0.885} & \textbf{0.120} & \textbf{0.886}

\\
\bottomrule
\end{tabular}
\label{tab:sup_cross_view_loss}
\end{table*}

\section{Visualization of Synthetic Underwater Rendering}
\label{appendix:render}

\begin{figure*}[t]
  \centering
  \includegraphics[width=0.95\linewidth]{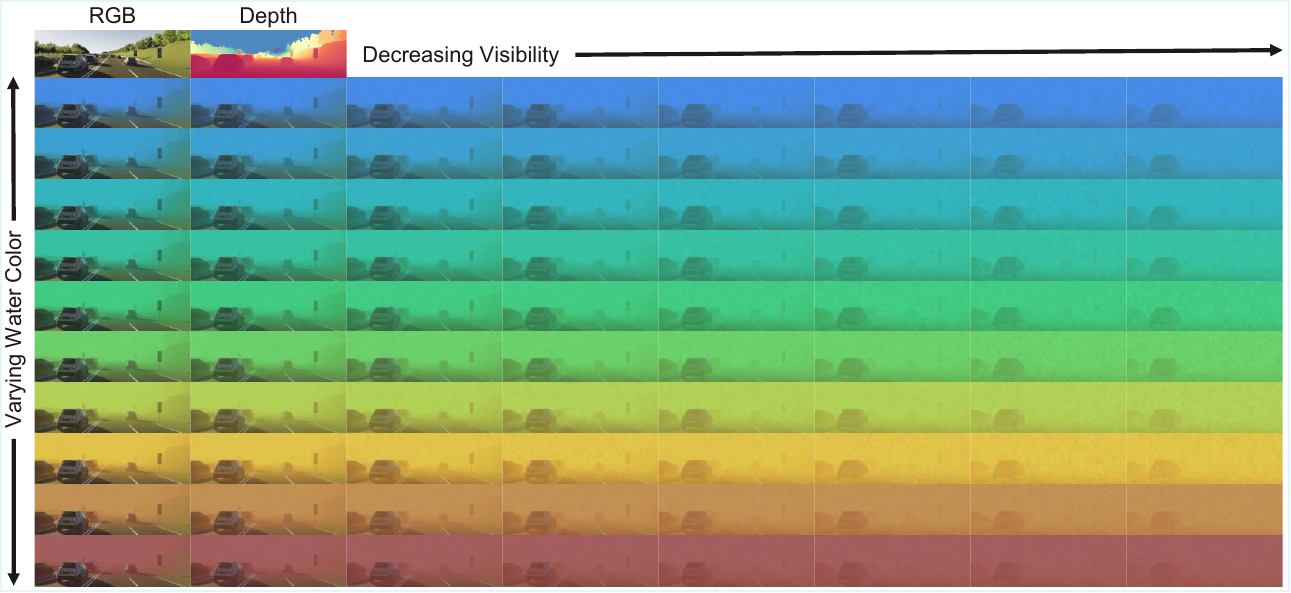}
\caption{\textbf{Synthetic underwater rendering examples.}
Starting from an on-land RGB image and depth map, we generate underwater images using our physics-based rendering model. The vertical axis varies the simulated water color, representing different underwater environments, while the horizontal axis shows decreasing visibility caused by increasing attenuation and scattering. }
\label{figs:sup_syn_water}  
\end{figure*}

To better illustrate the behavior of our physics-based underwater rendering process, we visualize several examples generated under different parameter settings in~\cref{figs:sup_syn_water}. The synthesized images encourage the model to learn the relationship between geometric structure and underwater light degradation.

\section{More Understanding of Our \methodname{} and \datasetname{}}
\label{appendix:understanding}

In this section, we present additional analyses to better understand the behavior of \methodname{}. 
First, we report condition-wise results on \datasetname{} to evaluate our method under different underwater environments.
Then, we compare our method with classical SfM pipelines to illustrate the difficulty of reliable reconstruction in underwater environments.

\subsection{Condition-wise Results on \datasetname{}}

We also report condition-wise results on \datasetname{} in~\cref{tab:water3d_summary}.
The scenes are grouped into shallow sea, deep sea, lake, and river environments according to the dominant capture condition.
\methodname{} consistently achieves the lowest overall mean and median errors across all groups.
River scenes remain the most challenging because they often contain stronger turbidity, less stable texture, and narrower visible range.

\begin{table}[!htbp]
\centering

\caption{\textbf{Condition-wise reconstruction error on \datasetname{}.} Best in \textbf{bold}.}

\footnotesize
\setlength{\tabcolsep}{6pt}
\resizebox{0.95\columnwidth}!{
\renewcommand{\arraystretch}{0.9}
\begin{tabular}{@{}lcccc}
\toprule

\multirow{3}{*}{\vspace{+0.5em}\textbf{Methods}} 
& \multicolumn{4}{c}{\textbf{Overall Mean / Median Error} $\downarrow$} \\
\cmidrule(lr){2-5}
& \textbf{Shallow Sea} 
& \textbf{Deep Sea} 
& \textbf{Lake} 
& \textbf{River} \\
\midrule

DA3~\tiny{(ICLR\textquotesingle26)} 
& 0.852 / 0.254
& 0.924 / 0.167
& 0.863 / 0.056
& 3.645 / 3.033
\\

VGGT~\tiny{(CVPR\textquotesingle25)} 
& 0.857 / 0.226
& 1.690 / 0.223
& 0.877 / 0.079
& 3.650 / 2.440
\\

\methodname{}~\tiny{(ours)} 
& \textbf{0.489} / \textbf{0.177}
& \textbf{0.283} / \textbf{0.101}
& \textbf{0.084} / \textbf{0.046}
& \textbf{0.703} / \textbf{0.513}
\\

\bottomrule
\end{tabular}
}
\label{tab:water3d_summary}
\end{table}

\subsection{Comparison with Optimization-based Underwater Pipelines}

To address whether the improvement holds beyond feed-forward baselines, we further compare with WaterSplatting~\cite{li2025watersplatting} and COLMAP-based SfM/MVS pipelines on FLSea Stereo in~\cref{tab:3dgs}.
For COLMAP, we test SuperPoint+LightGlue (SPLG) and MINIMA matchers.
The COLMAP variants can obtain low error on successfully reconstructed scenes, especially with MINIMA, but they fail on most 10-view testing sequences and therefore lack robustness under sparse underwater inputs.
WaterSplatting also suffers from inaccurate initialization and sparse-view reconstruction failures.
In contrast, \methodname{} produces valid predictions for all scenes and achieves the best overall robustness.

\begin{table}[!htbp]
\centering

\caption{\textbf{Comparison with underwater-specific and classical reconstruction pipelines on FLSea Stereo.} We evaluate multi-view depth under the same 10-view setting as the main paper.}
\footnotesize
\setlength{\tabcolsep}{6pt}
\renewcommand{\arraystretch}{0.9}
\begin{tabular}{@{}lcccc}
\toprule

Methods

&  Rel$\downarrow$  & $\delta1\uparrow$ 
& RMSE$\downarrow$ & Failure rate$\downarrow$
\\

\midrule
\methodname{} (ours) &  0.119

& 0.885

& 0.720

& 0/152
 \\ 
 WaterSplatting~\tiny{(3DV\textquotesingle25)} & 0.427

 & 0.415

 & 1.476

 & 99/152
\\

COLMAP+SPLG & 0.129

& 0.847

& 0.584

& 91/152
\\
COLMAP+MINIMA & 0.105

& 0.896

& 0.456

& 99/152
\\

\bottomrule
\end{tabular}
\label{tab:3dgs}
\end{table}

\subsection{Additional Ablations}

\cref{tab:strong_aug} further isolates the effect of sequence-level strong augmentation.
Applying strong augmentation to synthetic data alone does not improve over synthetic-only supervised training, indicating that the augmentation mainly benefits the semi-supervised real-video branch where viewpoint diversity is limited.
To verify that our adaptation strategy is not tied to VGGT, we also apply the same framework to $\pi3$~\cite{wang2025pi}.
As shown in~\cref{tab:pi3}, the adapted model improves both Sea-thru and FLSea Stereo, suggesting that the proposed cross-domain semi-supervised training can transfer to other feed-forward geometry backbones.

\begin{table}[!htbp]
\centering
\caption{\textbf{Additional ablation on synthetic data and strong augmentation on FLSea Stereo.} We report multi-view depth metrics to isolate the effect of synthetic underwater data, sequence-level strong augmentation (SA), and unlabeled real videos.}
\footnotesize
\setlength{\tabcolsep}{6pt}
\renewcommand{\arraystretch}{0.8}
\begin{tabular}{@{}lcccc}
\toprule
\textbf{Methods} & VGGT & +syn & +syn+SA & +syn+real+SA \\
\midrule
Rel$\downarrow$      & 0.136 & 0.131 & 0.132 & 0.126 \\
$\delta1\uparrow$     & 0.851 & 0.868 & 0.864 & 0.873 \\
\bottomrule
\end{tabular}
\label{tab:strong_aug}
\end{table}

\begin{table}[!htbp]
\centering
\caption{\textbf{Applying the proposed adaptation framework to $\pi3$~\cite{wang2025pi}.} We replace VGGT with $\pi3$ as the base feed-forward geometry model and evaluate multi-view depth on Sea-thru and FLSea Stereo.
}

\footnotesize
\setlength{\tabcolsep}{6pt}
\renewcommand{\arraystretch}{0.9}
\begin{tabular}{@{}lccc|ccc}
\toprule
\multirow{3}{*}{\vspace{+0.5em}\textbf{Methods}} 
 & \multicolumn{3}{c}{\textbf{Sea-thru}} 
 & \multicolumn{3}{c}{\textbf{FLSea Stereo}} 
\\
\cmidrule(lr){2-4} \cmidrule(lr){5-7}
&  Rel$\downarrow$  & $\delta1\uparrow$  
& RMSE$\downarrow$ & 
Rel$\downarrow$  & $\delta1\uparrow$ &

 RMSE$\downarrow$ 
\\

\midrule
$\pi3$~\tiny{(ICLR\textquotesingle26)}   &   0.185 & 0.909 & 0.358

& 0.139 & 0.856 & 0.837

\\
\methodname{} ($\pi3$ as base model)  & 0.169 & 0.933 & 0.310 &  0.117 & 0.899 & 0.763

 \\

\bottomrule
\end{tabular}
\label{tab:pi3}
\end{table}

\begin{figure*}[t]
  \centering
  \includegraphics[width=0.95\linewidth]{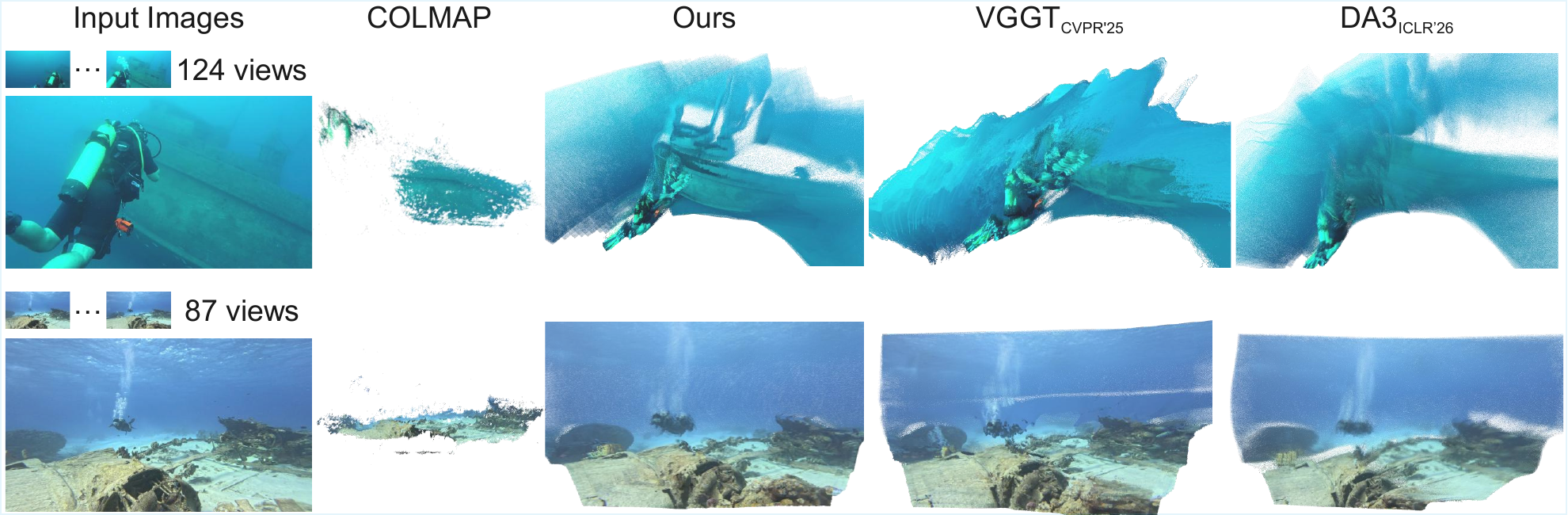}

\caption{
\textbf{Examples of challenging scenes from \datasetname{}.}
During the construction of the dataset, we use an improved COLMAP~\cite{schonberger2016structure} pipeline to reconstruct point clouds from raw underwater videos. 
However, many scenes still lead to unstable or degenerate reconstructions due to scattering effects, low texture, and limited camera motion. 
For comparison, we also show the results of \methodname{}, VGGT~\cite{wang2025vggt}, and DA3~\cite{lin2025depth}. 
While COLMAP fails to produce reliable reconstructions in these cases, our method is able to generate more coherent and geometrically consistent point clouds.
}
\label{figs:colmap_failure}  
\end{figure*}

\subsection{\methodname{} v.s. Colmap}

We analyze the robustness of different reconstruction approaches in challenging underwater environments. 
During the construction of \datasetname{}, we collected nearly 100 underwater video scenes and attempted to reconstruct them using an improved COLMAP~\cite{schonberger2016structure} pipeline to obtain ground-truth geometry. However, despite careful tuning and dense multi-view inputs, reliable reconstructions could only be obtained for 42 scenes after manual inspection, which were finally retained in the dataset. Several representative examples are shown in~\cref{figs:colmap_failure}.

The examples highlight the difficulty of applying classical SfM pipelines to underwater imagery. 
In the upper example of~\cref{figs:colmap_failure}, COLMAP reconstructs only a small portion of the ship hull surface while the remaining structure becomes fragmented with scattered outliers. 
Moreover, the diver’s oxygen tank visible in the upper-left corner is incorrectly reconstructed behind the ship hull, indicating significant geometric inconsistency. Another example is shown in the lower row. Although the images appear visually clear, the camera motion in the sequence is limited. 
Under such a small camera baseline, COLMAP produces a degenerate reconstruction where the recovered point cloud collapses into an almost planar structure and fails to recover the true scene geometry.

In contrast, learning-based approaches show significantly stronger robustness in these scenarios. 
Even when COLMAP fails to produce valid reconstructions, our method can still generate coherent and structurally consistent point clouds that better reflect the underlying scene geometry.

\section{Using v.s. Not Using underwater Image enhancement}
\label{appendix:uie}
Many recent underwater vision works~\cite{huang2023contrastive,liu2025toward} focus on restoring underwater images to resemble terrestrial scenes. 
Such restoration aims to reduce water-induced interference and provide higher-quality inputs for downstream tasks. Therefore, we additionally investigate the role of underwater image enhancement (UIE) in 3D reconstruction.
We further evaluate our method using the latest underwater image restoration approaches together with models trained on different scenes, as shown in~\cref{tab:image_enhence}.

However, this two-stage pipeline does not necessarily benefit 3D reconstruction. Most UIE methods are designed to improve visual clarity rather than geometric consistency, and they are typically trained independently of downstream geometric tasks. As a result, the enhanced images may not provide reliable cues for multi-view geometry estimation. Instead of explicitly removing water effects, we simulate underwater conditions by adding water effects to terrestrial images during training.
This strategy allows the model to implicitly learn water removal while understanding underwater geometry.
\begin{table}
\centering
\caption{\textbf{Performance change with underwater image enhancement.} We report Absolute Relative Error (Rel$\downarrow$) for multi-view depth estimation. Applying UIE as a preprocessing step does not lead to clear improvements in geometric estimation.}

\footnotesize
\setlength{\tabcolsep}{6pt}
\resizebox{0.95\columnwidth}!{
\begin{tabular}{@{}lccc|ccc}
\toprule
\multirow{3}{*}{\vspace{+0.5em}\textbf{Methods}} &
\multicolumn{3}{c}{\textbf{Sea-thru\cite{akkaynak2019sea}}~(Rel$\downarrow$)}
&
\multicolumn{3}{c}{\textbf{FLSea Stereo\cite{randall2023flsea}}~(Rel$\downarrow$)} \\
\cmidrule(lr){2-4} \cmidrule(lr){5-7}
&   VGGT & MapAnything & DA3 &  VGGT & MapAnything & DA3 
\\

\midrule
Original & 0.190 & 0.216 & 0.187 & 0.137 & 0.146 & 0.141
 \\
 
Semi-UIR~\cite{huang2023contrastive}~\tiny{CVPR\textquotesingle23} & 0.201 & 0.218 & 0.195 & 0.136 & 0.148 & 0.147
  \\
  
PSPL~\cite{liu2025toward}~\tiny{TIP\textquotesingle25} & 0.209 & 0.230 & 0.207 & 0.160 & 0.151 & 0.149
 \\

\bottomrule
\end{tabular}
}
\label{tab:image_enhence}
\end{table}

\section{Additional Visualization}
\label{appendix:vis}

\begin{figure*}[t]
  \centering
  \includegraphics[width=0.95\linewidth]{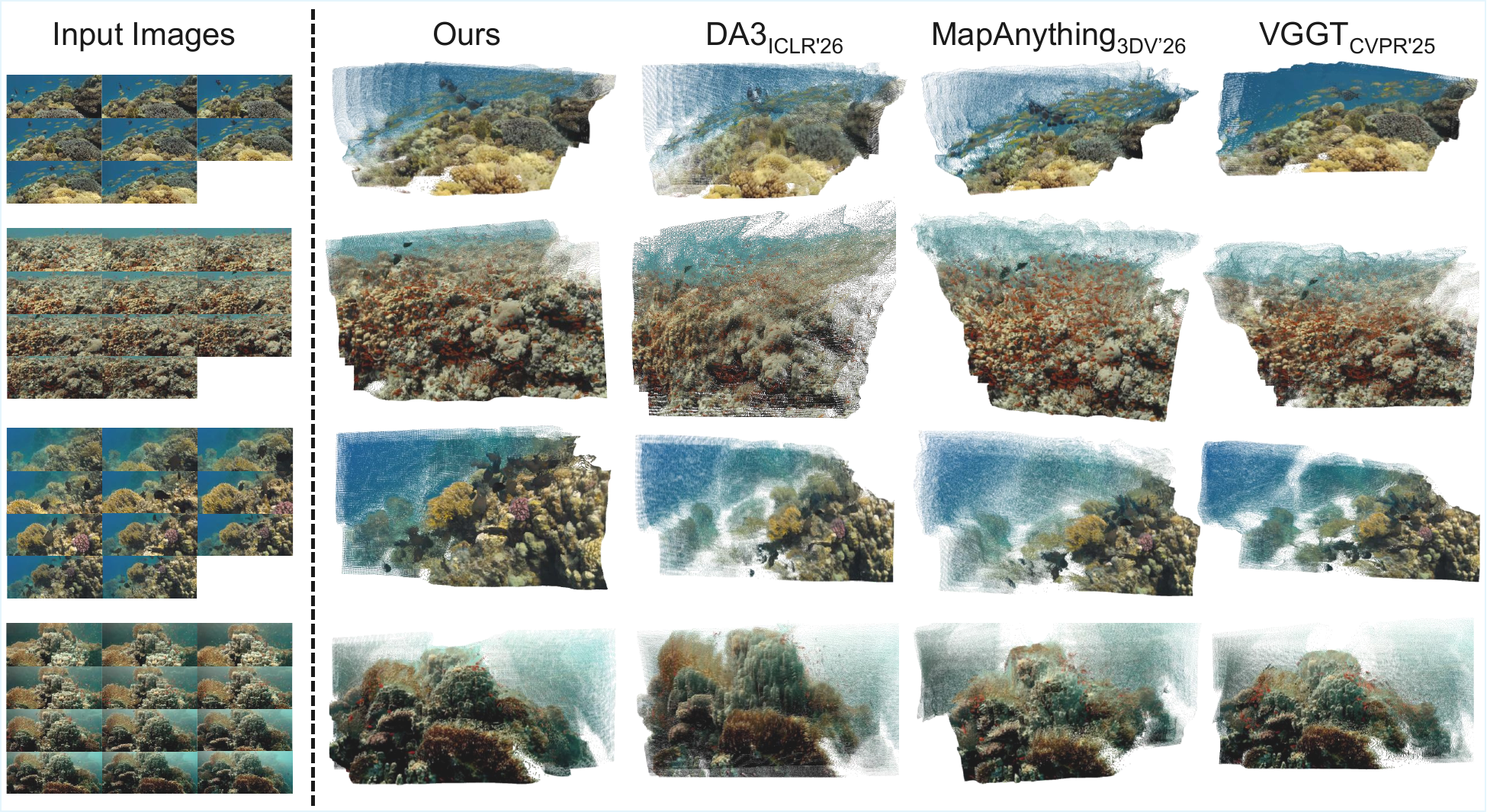}
    \caption{\textbf{Qualitative comparison of in-the-wild multi-view 3D reconstruction.}
We compare \methodname{} with DA3~\cite{lin2025depth}, MapAnything~\cite{keetha2025mapanything}, and VGGT~\cite{wang2025vggt} on challenging underwater scenes. 
Our method produces more coherent and complete 3D structures, while competing methods suffer from fragmented geometry or missing structures.}
\label{fig:sup_point_in_the_wild}  
\end{figure*}
\begin{figure*}[htbp]
  \centering
  \includegraphics[width=0.95\linewidth]{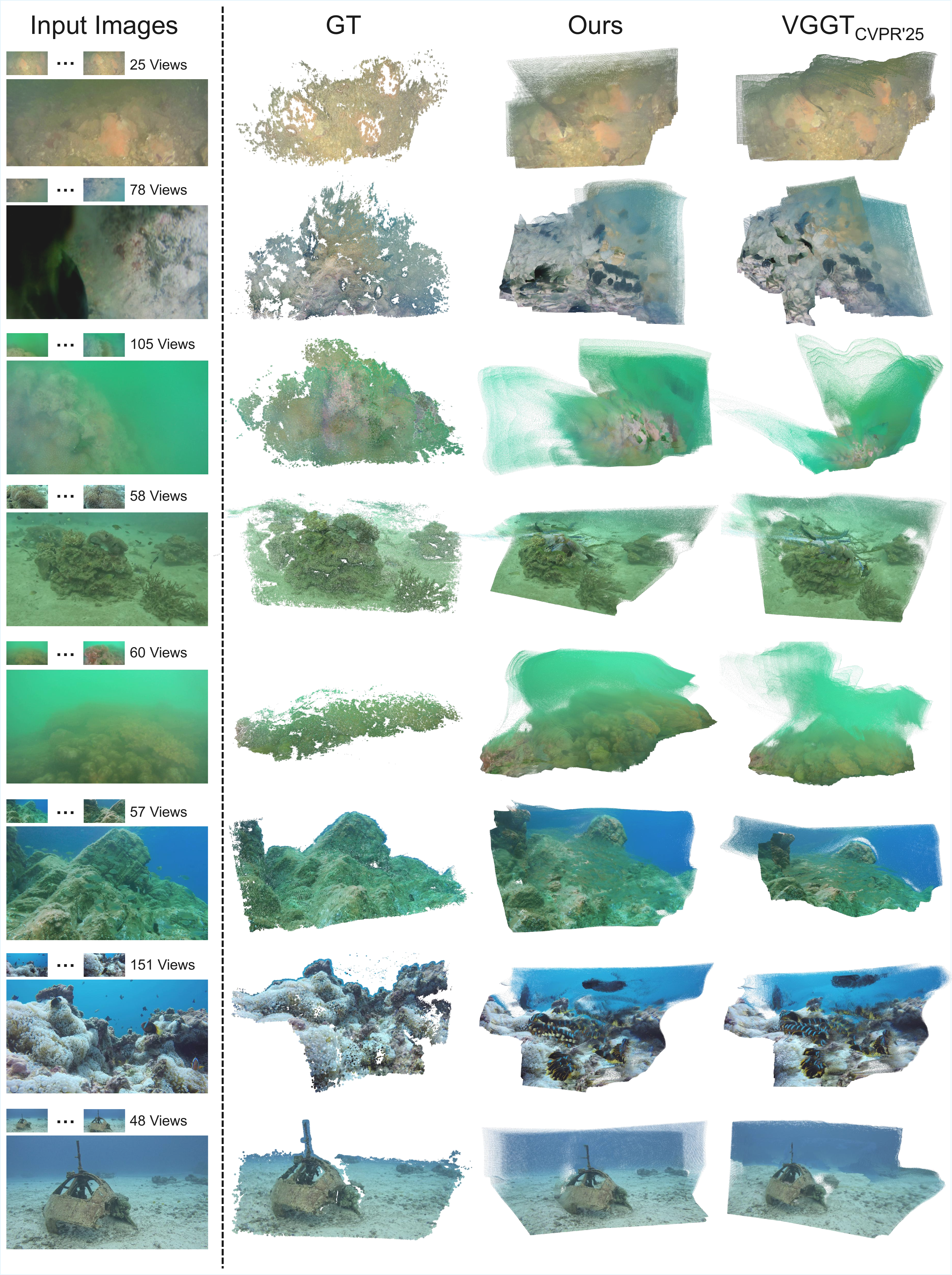}
\caption{\textbf{Qualitative results on diverse underwater scenes from \datasetname{}.}
The examples illustrate the diversity of underwater environments in our dataset. 
Compared with VGGT~\cite{wang2025vggt}, \methodname{} produces more complete and coherent reconstructions, even recovering structures that are missing in the COLMAP-based ground truth.}
\label{figs:water_result}  
\end{figure*}
\begin{figure*}
  \centering
  \includegraphics[width=0.95\linewidth]{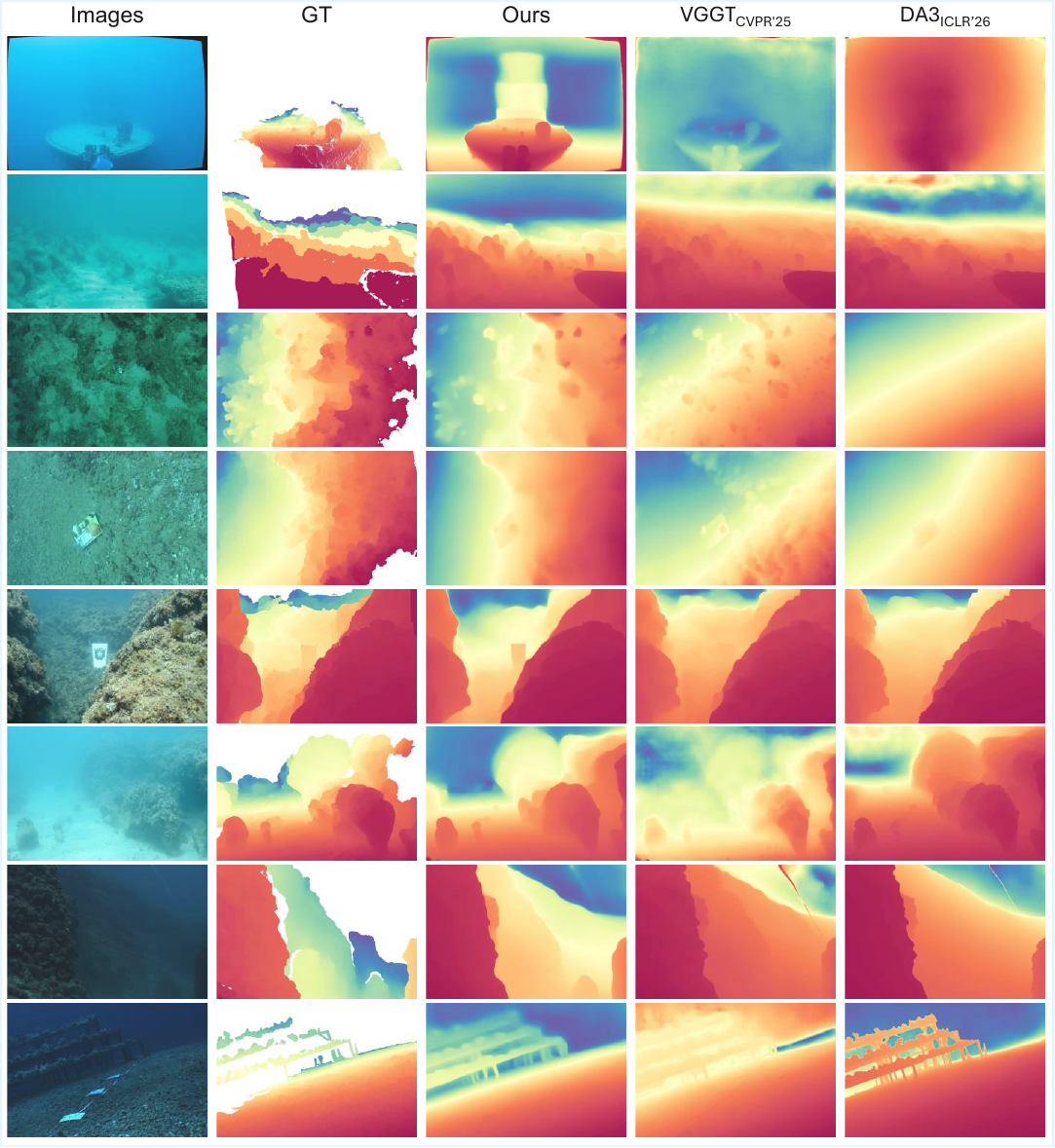}
\caption{\textbf{Qualitative results on monocular depth estimation using \methodname{}, DA3~\cite{lin2025depth} and VGGT~\cite{wang2025vggt}.}
Rows 1--2: SQUID~\cite{berman2020underwater}; rows 3--4: SeaThru~\cite{akkaynak2019sea} (brightness increased for visualization only); rows 5--6: FLSea Stereo~\cite{randall2023flsea}; rows 7--8: FLSea VI~\cite{randall2023flsea}.}
\label{figs:sup_depth_result}  
\end{figure*}

We provide additional geometric prediction results for in-the-wild multi-view inputs in~\cref{fig:sup_point_in_the_wild}, multi-view inputs on our \datasetname{} in~\cref{figs:water_result}, and monocular inputs in~\cref{figs:sup_depth_result}.It can be seen that our model achieves robust and high-quality geometric predictions across several scenarios.

\end{document}